\documentclass[letterpaper, 10 pt, conference]{ieeeconf}  

\IEEEoverridecommandlockouts                              

\usepackage{booktabs} 
\usepackage{graphicx}
\usepackage{amsmath}
\usepackage{amsfonts}
\usepackage{xspace}
\usepackage{multirow}
\usepackage{twemojis}
\usepackage{xcolor}
\usepackage{soul}
\usepackage{wrapfig}
\usepackage{bbding}
\usepackage{pifont}
\usepackage{subcaption}
\usepackage{makecell}
\usepackage{listings}
\usepackage{hyperref}
\usepackage{cuted}
\usepackage{capt-of}
\usepackage[export]{adjustbox}   
\usepackage{cite}
\usepackage{tabularx}
\newcolumntype{Y}[1]{>{\raggedright\arraybackslash\hsize=#1\hsize}X}

\def\eg{\emph{e.g.,}\xspace} 

\def\ie{\emph{i.e.,}\xspace}

\newcommand{\pizero}{$\pi_{0}$\xspace}

\newcommand{\pizerofivebold}{\ensuremath{\boldsymbol{\pi_{0.5}}}\xspace}
\newcommand{\otter}{OTTER\xspace}
\newcommand{\pizerofive}{$\pi_{0.5}$\xspace}
\newcommand{\ourvla}{w$^{2}$VLA\xspace}

\definecolor{wwvlarow}{HTML}{E8F4F5}
\definecolor{lightBlue}{HTML}{E6F0FA}

\definecolor{lightGreen}{HTML}{E6FAE6}

\definecolor{green_highlight}{HTML}{4a7a55}

\definecolor{yellow_highlight}{HTML}{8a6b2f}

\definecolor{red_highlight}{HTML}{8c4a41}

\definecolor{blue_highlight}{HTML}{46628f}

\usepackage[table]{xcolor}

\definecolor{DeltaLight}{RGB}{255,255,255} 
\definecolor{DeltaDark}{RGB}{200,120,110}  

\newcommand{\dcell}[1]{\cellcolor{DeltaDark!#1!DeltaLight}\textcolor{black}{$\downarrow$\,#1}}

\definecolor{deltaBg}{RGB}{220,230,255} 

\title{\LARGE \bf
Decoupling the \textit{Declarative} from the \textit{Procedural}\\in Vision-Language-Action Models
}

\author{\ Nikolaos Tsagkas$^{1}$,\ Andreas Sochopoulos$^1$,\ Chris Xiaoxuan Lu$^{2}$,\ Oisin Mac Aodha$^{1}$, Alexandros Kouris$^{3}$\\ 
  $^1$University of Edinburgh, \ $^2$UCL, \ $^{3}$Samsung AI Center - Cambridge, UK\\
}

\makeatletter
\def\bstctlcite{\@ifnextchar[{\@bstctlcite}{\@bstctlcite[@auxout]}}
\def\@bstctlcite[#1]#2{\@bsphack
  \@for\@citeb:=#2\do{%
    \edef\@citeb{\expandafter\@firstofone\@citeb}%
    \if@filesw\immediate\write\csname #1\endcsname{\string\citation{\@citeb}}\fi}%
  \@esphack}
\makeatother

\begin{document}
\bstctlcite{BSTcontrol} 
\maketitle
\thispagestyle{empty}
\pagestyle{empty}

\begin{strip}
  \vspace{-38pt} 
  \centering
  \makebox[\textwidth][c]{\includegraphics[width=0.98\textwidth]{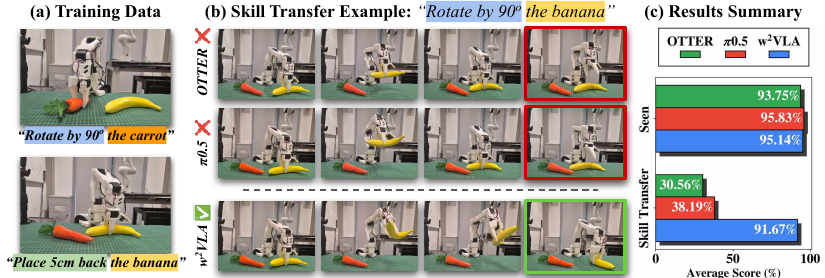}}
  
  \par 
  \vspace{5pt} 
  
  \captionof{figure}{\textbf{Skill transfer example}. (a) Three VLAs (\ie \pizerofive, \otter, and our \ourvla) are trained on a dataset of two \textit{(skill, object)} pairs: \textit{(rotate 90$^o$, carrot \twemoji{carrot})} and \mbox{\textit{(place 5cm back, banana \twemoji{banana})}}. (b) We evaluate the three VLAs in a skill transfer scenario \mbox{\textit{(rotate 90$^o$, banana \twemoji{banana}})}. Unlike our \ourvla, \pizerofive and \otter fail to transfer the learned skill to the other object. (c) Summary of our main experimental results: all three VLAs obtain almost perfect performance when deployed over seen \textit{(skill, object)} pairs, however, only \ourvla manages to reliably perform skill transfer.}
  \vspace{-10pt} 
  \label{fig:teaser_overview}
\end{strip}

\begin{abstract}
Deploying generalist robotic agents in the real world requires transferable skills. 
Specifically, a policy trained to clone a behavior from object-specific demonstrations must generalize beyond that object, otherwise data collection requirements become intractable.
Recently, fine-tuning of pre-trained billion-parameter Vision-Language Models (VLMs), initially on large-scale robot datasets and then on fewer scenario-specific demonstrations, has emerged as the predominant paradigm for designing Vision-Language-Action (VLA) models. 
While these policies achieve state-of-the-art manipulation performance in-distribution, they remain brittle to minor spatial, semantic, and task variations.  
In this work, we address the inability of current models to decouple the declarative (\ie concepts and entity semantics) from the procedural knowledge (\ie how to do something) encoded in their parameters, which is a fundamental bottleneck for zero-shot skill transfer to novel objects.
To address this, we propose \ourvla, a new VLA model with restructured information flow. 
Rather than feeding all multimodal tokens from the VLM encoder into a large, opaque transformer-based action expert, our approach  modulates the robot state sequence with visual, spatial, and skill information in a compositional and interpretable manner. 
Unlike popular, state-of-the-art VLAs, we show that our modular approach successfully decouples knowledge representations, enabling robust behavior cloning and unprecedented zero-shot skill transfer capabilities across dissimilar, unseen objects.
\textbf{Project Page}:~\texttt{\href{https://tsagkas.github.io/w2vla}{https://tsagkas.github.io/w2vla}}.\newline
\end{abstract}

\begin{figure*}[t]
    \centering
    \vspace{5pt}
    \includegraphics[width=0.95\linewidth]{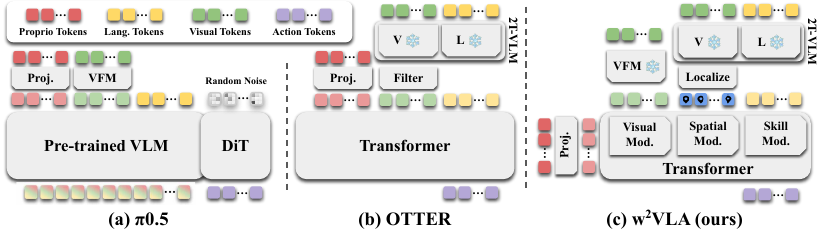}
    \vspace{-3pt}
    \caption{We compare two popular VLA paradigms against our \ourvla. \textbf{(a)} \pizerofivebold maps all input tokens into the VLM's input space, which gets fine-tuned during training, followed by a Diffusion Transformer (DiT) action expert. \textbf{(b) \otter} extracts visual and language tokens from a frozen two-tower VLM (2T-VLM) and performs a textual-aware filtering of the visual features before concatenating the multimodal tokens into a single input to a causal transformer action expert. Instead of treating all multimodal tokens as a unified sequence, our \textbf{(c) \ourvla} sequentially modulates the proprioception tokens with visual, spatial, and skill tokens extracted from frozen 2T-VLM and Vision Foundation Model (VFM) backbones. Only \ourvla demonstrates skill transfer capabilities.}
    \vspace{-13pt}
    \label{fig:vla_comparison}
\end{figure*}

\section{Introduction}
\label{sec:intro}
In the pursuit of generalist, autonomous robot agents, the paradigm for training policies via imitation learning (IL) has shifted toward fine-tuning deep, billion-parameter Vision-Language Models (VLMs) (\eg PaliGemma~\cite{beyer2024paligemmaversatile3bvlm}, Qwen-VL~\cite{wang2024qwen2}). 
This is motivated by their impressive generalization capabilities in traditional vision tasks. 
This trend, however, is rooted in a deeper assumption: that foundation models in robotics will emerge, much like in the vision and language domains, through the sheer scaling of both model parameters and datasets. 
Consequently, the recent proliferation of large-scale robotics datasets (\eg OXE~\cite{o2024open}, DROID~\cite{khazatsky2024droid}) has been instrumental in accelerating this trajectory. 
These architectures, termed Vision-Language-Action (VLA) models~\cite{pmlr-v270-kim25c,bjorck2025gr00t, black2024pi_0,black2025pi05,intelligence2025pi06vlalearnsexperience,shukor2025smolvla,goyal2025vla0buildingstateoftheartvlas}, have indeed exhibited remarkable manipulation capabilities, even when fine-tuned with just a few scenario-specific demonstrations (tailored to the target task, object, scene, and embodiment), while achieving state-of-the-art performance on popular benchmarks (\eg LIBERO~\cite{liu2023libero}, Robotwin~\cite{chen2025robotwin}).

However, despite the aforementioned success, recent work has revealed that modern VLAs lack the robust generalization capabilities anticipated when deploying strong VLM backbones~\cite{fei25libero-plus,zhou2025liberopro,huang2025otter,yang2026robolabhighfidelitysimulationbenchmark}. 
Minor environmental variations, such as altering object arrangements in the scene or modifying fundamental attributes (\eg object color), can cause the trained policy to fail completely. 
This raises a critical question: \textit{what is the utility of computationally expensive pre-training and fine-tuning pipelines if the resulting policies remain susceptible to the same trivial perturbations that plague traditional IL approaches?} 
Recent literature has attempted to address these vulnerabilities by freezing the VLM entirely (\eg \otter~\cite{huang2025otter}) or partially (\eg OpenVLA$^+$~\cite{grover2025enhancinggeneralizationvisionlanguageactionmodels}). 
We argue that, conceptually, this is a highly promising direction. 
VLMs excel at encoding \textit{declarative knowledge}, \ie objects' identities, spatial properties and visual semantics, which should remain fundamentally decoupled from \textit{procedural knowledge}, \ie the mechanics of executing a skill.
For instance, learning to grasp a mug should not require the model to relearn visually recognizing a mug. 

Nevertheless, both VLAs with fine-tuned backbones and those with frozen VLMs lack a capability crucial for generalist agents: \textit{zero-shot skill transfer across distinct objects}.
While humans readily extrapolate a skill from one object to another (\eg a child who learns to hold a hammer can intuitively adapt that grasp to a wrench), this remains a profound challenge even for state-of-the-art VLAs (Fig.~\ref{fig:teaser_overview}).
We attribute this to their opaque action experts, which process all visual, language, and proprioceptive tokens as a single unified sequence (Fig.~\ref{fig:vla_comparison}(a,b)).
Given the few demonstrations available for scenario-specific fine-tuning, this encourages overfitting to spurious skill-object correlations, which scaling data and parameters alone cannot tractably resolve.

We instead follow the cognitive principle of decoupling perception from action, loosely analogous to the division of labor in the human visual system~\cite{GOODALE199220,goodale1991neurological,milner2006visual}, and separate a task's \textit{spatial grounding} (the \textit{where}) from its \textit{motor intent} (the \textit{what}).
The \textit{where} is declarative, object-centric knowledge already encoded by the VLM, while the \textit{what} is procedural knowledge that must be learned from demonstrations without interfering with these priors.
To this end, we cast the sequence transduction of transformers~\cite{attention_is_all_you_need} as a mapping from a history of proprioceptive robot states to future actions, and sequentially modulate these states with visual, \textit{where}, and \textit{what} information (Fig.~\ref{fig:vla_comparison}(c)).
At each timestep, the transformer thus \textit{transforms} its hidden states by \textit{how much they need to change} to guide the robot to a \textit{location} and clone a \textit{behavior}, unbinding the skill from the object it was demonstrated on.
CLIPort~\cite{shridhar2021cliport} draws on the same two-stream analogy, but its semantic (\textit{what}) and spatial (\textit{where}) pathways are both perceptual, jointly predicting stateless, discrete $SE(2)$ pick-and-place affordances.
Our \textit{what} instead encodes procedural skill knowledge, decoupled from perception within the continuous, temporally-conditioned action policies of modern VLAs.

We instantiate this idea in \ourvla (\textit{where-what-VLA}), an end-to-end trainable VLA that, to the best of our knowledge, is the first to achieve zero-shot skill transfer to unseen objects.
Akin to \otter, it keeps a two-tower VLM (\ie MetaCLIP2~\cite{chuang2025metaclip2worldwide}) frozen, so its pre-trained representations remain undiluted by the low-data regime of IL.
Through real-world comparisons, we show that \ourvla (\textit{i}) enables compositional generalization, transferring skills between demonstrated objects where strong baselines collapse; (\textit{ii}) maintains this robust execution on completely unseen objects; and (\textit{iii}) matches the in-domain behavior cloning performance of VLAs with fine-tuned (\eg \pizerofive) and frozen (\eg \otter) backbones.
Finally, a single \ourvla trained on all 40 LIBERO tasks shows that this structure extends to standard multi-task benchmarks.
\vspace{-5pt}

\begin{figure}
    \vspace{5pt}
    \centering
    \includegraphics[width=0.98\linewidth]{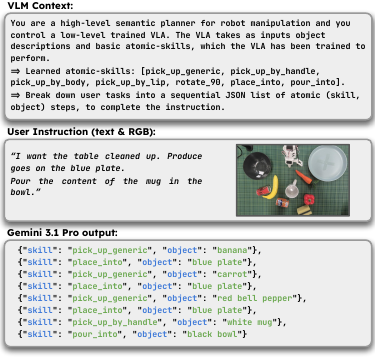}
    \vspace{-7pt}
    \caption{\textbf{Command interface}.
    A VLM planner, prompted zero-shot with \ourvla's atomic skills, decomposes an abstract instruction and an RGB observation into a sequence of \textit{(skill, object)} commands that our policy consumes.
    }
    \label{fig:command_interface}
    \vspace{-16pt}
\end{figure}

\section{The \texorpdfstring{w$^2$VLA}{w²VLA} Model}
\label{sec:methodology}
\begin{figure*}[t]
    \vspace{5pt}
    \centering
    \includegraphics[width=0.95\textwidth]{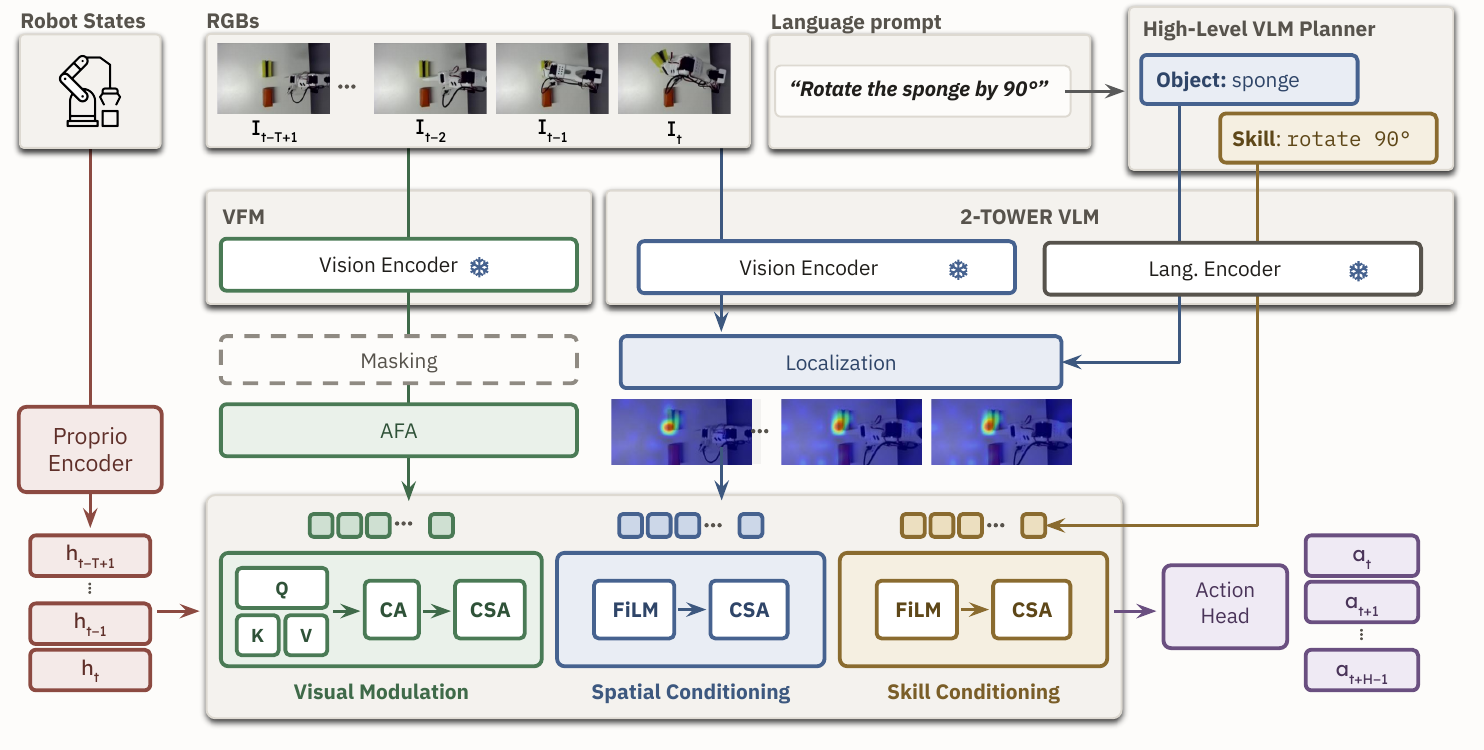}
    \vspace{-10pt}
    \caption{\textbf{w$^2$VLA architecture}. Our policy sequentially modulates a sequence of \textbf{\textcolor{red_highlight}{hidden robot state tokens}} with the decoupled \textit{where} (\ie location of interest) and \textit{what} (\ie skill to be executed) information of the task. \textbf{\textcolor{green_highlight}{Visual tokens}} are extracted from a VFM encoder to inform the hidden states of relevant visual cues in the scene. \textbf{\textcolor{blue_highlight}{Where (spatial) tokens}} are computed via attention heatmaps that localize the object of interest in space. \textbf{\textcolor{yellow_highlight}{What (skill) tokens}} represent the skill to be executed.}
    \label{fig:vla_architecture}
    \vspace{-15pt}
\end{figure*}

We begin by formulating the skill transfer problem in the context of VLAs (Sec.~\ref{ssec:problem}), and describe the architectural components (Sec.~\ref{ssec:architecture}) and training mechanism  (Sec.~\ref{ssec:training}) of our proposed \ourvla (where-what-VLA) model. 
\ourvla is designed to adopt a restructured information flow that decouples declarative from procedural knowledge. 
At the core of this new architecture lies a novel action expert transformer, equipped with tailored conditioning blocks that sequentially modulate the internal representation with \textit{visual}, \textit{spatial}, and \textit{skill} information, extracted from pre-trained foundation models. 
This isolation and progressive conditioning on different information components during the mapping process of robot states to actions, is  instrumental in enabling skill transfer beyond the demonstration training tasks. 
The proposed architecture is illustrated in Fig.~\ref{fig:vla_architecture}. 

\subsection{Problem Formulation}
\label{ssec:problem}
In order to study the skill transfer capabilities of VLAs we formulate a simple and targeted IL-based manipulation scenario. 
Specifically, we assume two distinct objects, $\text{o}_1$ and $\text{o}_2$, with randomized initial positions within the robot workspace. 
Each object is coupled with a distinct skill, $\text{s}_a$ or $\text{s}_b$, that the robot performs on it (\eg rotating a carrot or moving back a banana, as in Fig.~\ref{fig:teaser_overview}) creating two primary skill-object pairs: $\langle\text{s}_a, \text{o}_1\rangle$ and $\langle\text{s}_b, \text{o}_2\rangle$, each linked with a language instruction describing the target object and skill to be executed. We collect an equal number of expert training demonstrations for each $\langle\text{s}, \text{o}\rangle$ pair, for varying object poses.

Under this setting, we consider skill transfer to be the successful execution of a skill to new objects, by interchanging the skill-object pairs of the demonstrations, \ie $\langle\text{s}_a, \text{o}_2\rangle$ and $\langle\text{s}_b, \text{o}_1\rangle$, in the input language instruction.
Working with two skill-object pairs allows us to evaluate the utilization of both the \textit{declarative} knowledge of the VLM (\ie if the correct object is selected) and that of the \textit{procedural} knowledge encoded in the policy (\ie generating the actions of the instructed skill).
However, \ourvla is a fully general VLA and is capable of extending skill transfer to completely unseen objects, \ie $\langle\text{s}_a, \text{o}_2\rangle,\langle\text{s}_a, \text{o}_3\rangle, \ldots,\langle\text{s}_a, \text{o}_K \rangle$ or $\langle\text{s}_b, \text{o}_1 \rangle,\langle\text{s}_b, \text{o}_3 \rangle, \ldots,\langle\text{s}_b, \text{o}_K \rangle$; see Sec.~\ref{ssec:robustness}.

\subsection{Model Architecture}
\label{ssec:architecture}

\noindent\textbf{Model Inputs}: At each timestep $t$, the model receives a temporal history of length $T$, comprising proprioceptive states $P=\{p_{t-T+1}, \dots, p_t\}$ and visual observations $I=\{I^c_{t-T+1}, \dots, I^c_t\}$ across $C$ cameras. 
Each image is sliced into $N$ patches and processed through two vision encoders with complementary strengths: a VFM extracts $D$-dimensional patch tokens $F \in \mathbb{R}^{T \times C \times N \times D}$ that provide robust spatial and semantic perception, while a VLM vision encoder extracts patch embeddings $V \in \mathbb{R}^{T \times C \times N \times D}$  that are strictly aligned with the text space. 

\noindent\textbf{Command Interface}:
\ourvla consumes an atomic command $c=(l_{skill}, l_{obj})$, comprising a \textit{skill} string and a \textit{target object} string.
This is the granularity at which low-level policies are already driven in hierarchical VLA systems~\cite{shi2025hi,torne2026memmultiscaleembodiedmemory,chen2026steerable}, and in particular by \pizerofive, whose high-level VLM predicts an atomic language subtask before its action expert is conditioned on it~\cite{black2025pi05}.
Relative to that prevailing paradigm we add exactly one named slot, in that the target object is stated explicitly rather than left implicit inside the subtask string.
We stress that this is a change of format and not a restriction of scope: Sec.~\ref{ssec:sim} passes all 40 LIBERO tasks through this interface without exception.
The two strings are independently encoded by the frozen VLM text encoder, yielding an object embedding $e_{obj} \in \mathbb{R}^D$ and a skill embedding $e_{skill} \in \mathbb{R}^{D}$.
In all experiments of Sec.~\ref{sec:experiments} the instruction is already atomic, so the split is trivial and \ourvla and all baselines receive the same raw instruction text.
For abstract, multi-step instructions, the command sequence is produced by an off-the-shelf VLM (\ie Gemini 3.1 Pro), prompted zero-shot with the closed set of atomic skills the policy was trained on and without any fine-tuning.
Fig.~\ref{fig:command_interface} shows this on a cluttered tabletop, where the planner grounds ``produce'' to the items actually present, routes each to the blue plate, and selects the handle-specific pick variant for the mug before pouring.
Task structures that appear to lie outside an atomic interface are reached by sequencing: ``put A into B'' becomes {\small\texttt{pick\_up}} A followed by {\small\texttt{place\_into}} B, referential targets (\eg ``the mug on the right'') are resolved into a referring expression at the language level, and contact-specific behaviors are exposed as dedicated skills (\eg {\small\texttt{pick\_up\_by\_handle}}) with the coarse heatmap acting only as a spatial guide.
Because the skill vocabulary is closed, a malformed skill is detectable before execution, whereas an incorrect object string degrades into a localization failure rather than a silent behavioral one.

\noindent\textbf{Encoding Robot States}: The proprioception sequence $P$ is first mapped via an MLP (\ie the \textit{Proprio Encoder}) into the policy's $D$-dimensional latent space.
Then, a temporal positional embedding is added to preserve sequential order, yielding the hidden robot states $H=\{h_{t-T+1}, \dots, h_t\}$. 
This sequence serves as the foundational representation which is fed as input to the \ourvla's action expert, where it will be subsequently modulated with multi-modal task context, guiding the prediction of next robot actions.

\noindent\textbf{Visual Modulation}: To capture correlations between the observed visual context and robot actions, the hidden robot states $H$ are first modulated with the dense VFM patch tokens $F$. 
Prior to that, we employ Attentive Feature Aggregation (AFA)~\cite{tsagkas2025attentivefeatureaggregationor,fu2024icrt,huang2025otter}, to filter out redundant background information. 
Using cross-attention with a set of $Q$ trainable queries, AFA compresses the visual features at each timestep $t$ into a concise set of task-relevant summary tokens $F'=\{f'_{\tau,q}\}$, where $\tau \in \{t-T+1, \dots, t\}$ and $q \in \{1, \dots, Q\}$ indexes the respective query. 
A cross-attention (CA) layer then integrates this visual context into the robot's hidden states by treating the summarized tokens $F'$ as keys and values, and the hidden states $H$ as queries. 
This module concludes with a Causal Self-Attention (CSA) block, allowing the visually-modulated tokens to temporally reason (and extrapolate) over the sequence history before propagating to the subsequent stages. 

\noindent\textbf{Conditioning Block}: To condition the base representation of hidden robot states on the task's \textit{where} (target object spatial location) and \textit{what} (skill to execute), we devise a conditioning block that sequentially modulates the action expert's hidden states on an external information signal. 
To achieve this, we adopt a Feature-wise Linear Modulation (FiLM)~\cite{film} based transformer block design, inspired by how instruction embeddings conditioned RT-1~\cite{brohan2023rt1roboticstransformerrealworld}. 
Specifically, an MLP projects the target conditioning vector $\sigma$ into scaling ($\gamma(\sigma)$) and shifting ($\beta(\sigma)$) parameters, which apply an affine transformation directly to each state token: $h'_\tau = FiLM(h_\tau, \sigma) =  \gamma(\sigma) \odot h_\tau + \beta(\sigma)$. 
This dynamically biases the tokens to reflect the precise skill intent or spatial target required at that exact moment. 
A subsequent CSA block models temporal dynamics, allowing each modulated token to attend to its updated history. 
Crucially, we employ a Pre-LayerNorm architecture~\cite{xiong2020layer} within the CSA and a residual connection around it. 
This allows the modulated features from the preceding FiLM block to bypass normalization, preventing a destructive standardization of the feature space, that would cancel out the injected $\gamma$ and $\beta$ modulation parameters: $h''_{\tau} = h'_{\tau} + \text{CSA}(\text{LayerNorm}(h'_{\tau}))$.

\begin{figure*}[t]
    \vspace{5pt}
    \centering
        \includegraphics[width=0.98\textwidth]{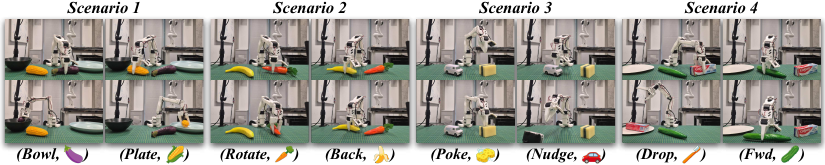}
    \caption{Visualization of all \textit{(skill, object)} pairs.}
    \label{fig:object_skill_pairs}
    \vspace{-18pt}
\end{figure*}

\noindent\textbf{Spatial Conditioning (\ie the \textit{where})}: Spatial conditioning is achieved by injecting a sequence of tokens $\Sigma_{\tau}^{spatial}=\{\sigma^{spatial}_{t-T+1}, \dots, \sigma^{spatial}_t\}$ into the first conditioning block, to ground the location of the target object in each visual observation to its corresponding hidden-space robot state token. 
We derive these conditioning tokens from spatial localization heatmaps $M_{\tau}=\{m_{t-T+1}, \dots, m_t\}$. 
Specifically, raw similarity scores are computed via the tensor product between the VLM's multi-camera visual patch embeddings and the textual object embedding: $S_t = V_t \cdot e_{obj}$, for each timestep. 
To transform these similarities into a precise spatial probability distribution, we apply a temperature-scaled Softmax over the spatial dimension independently for each camera: $m_t = \text{Softmax}(S_t / \zeta)$, where $\zeta = 0.05$ is a hyperparameter that controls the sharpness of the attention map. 
The resulting heatmaps $m_{t}$ are flattened and projected via an MLP into the transformer's hidden dimension to form the final conditioning tokens $\sigma^{spatial}_{t}$. 
By sequentially modulating the robot states with $\Sigma^{spatial}$, the policy is explicitly biased to focus on the target object's geometric region independently of the object's appearance and instructed task, successfully decoupling the spatial (\textit{where}) information from the  remaining components of skill execution. 

\noindent\textbf{Skill Conditioning (\ie the \textit{what})}: To condition the policy on the behavior intent of the task (e.g., ``pick'', ``push''), we utilize the skill embedding $e_{skill}$. 
As the semantic goal remains constant throughout the execution of a given primitive, $e_{skill}$ is projected via an MLP into the transformer's hidden dimension to yield a single skill vector per instruction: $\sigma^{skill} = \text{MLP}(e_{skill})$, 
that is broadcast within the skill conditioning block, across the temporal dimension $T$ of a rollout. 
By sequentially routing $\sigma^{skill}$ through this block, the policy maps the localized object representation, established by the preceding spatial module, to the specific motor trajectory dictated by the language instruction. 

\noindent\textbf{Action Head}:  The fully conditioned sequence of hidden robot states is processed by an MLP that predicts a chunk of future actions $\hat{A} = \{ a_{t+1},  a_{t+2}, \dots,  a_{t+L}\}$, where $L$ is the chunk length. 
Without loss of generality, we hereafter parametrize action predictions as end-effector pose deltas. 

\subsection{Model Training}
\label{ssec:training}
The proposed policy $\pi_{\theta}$ is optimized through imitation learning over a scenario-specific demonstration dataset $\mathcal{D}$, using a standard behavior cloning loss, as: $\mathbb{E}_{(p_t^i, I_t^i, l^i, a_t^i)\sim\mathcal{D}} [\mathcal{L}_1(a_t^i,\pi_{\theta}(p_t^i, I_t^i, l^i))]$. 
In order to facilitate the emergence of zero-shot skill transfer capabilities, even under a severely limited number of demonstrations, we complement the restructured information flow of our architecture with two key elements in the adopted training recipe. 
First, the foundation models employed for vision and language encoding (\ie VFM and VLM backbones) remain frozen throughout policy training. 
This ensures that their rich, pre-trained representations remain undiluted by the traditionally low-data regime of IL, maintaining their generalization capabilities that  enable the recognition and localization of novel objects beyond the provided demonstration data. 
Second, we observe that raw visual signals can hinder skill transfer by introducing strong appearance biases tied to specific skill-object pairs observed during training. 
To mitigate this, we apply a structured dropout approach during training, inspired by~\cite{he2022masked}, masking VFM patches entirely with a probability $q=0.5$. 
This encourages the model to densely rely on the rich spatial and skill information from the newly introduced where and what conditioning modules in order to predict the desired behavior, enforcing the decoupling between the visual appearance of the object and the skill that is executed. 

\begin{table*}[t]
\vspace{5pt}
\centering
\caption{\textbf{Experiment success scores (\%).} Each scenario tests two distinct skills, originally trained on specific objects. \textit{Seen} denotes evaluating the skill on its originally paired object, while \textit{Transfer} evaluating the skill on the alternative object within the same scenario. 
We denote with \textcolor{green}{\ding{52}} the cases where the model qualitatively exhibited the desired skill transfer properties in the majority of rollouts and with \textcolor{red}{\ding{56}} where it did not. 
All scenarios are visualized in Fig.~\ref{fig:object_skill_pairs}.
}
\setlength{\tabcolsep}{4pt}
\renewcommand{\arraystretch}{0.85}
\resizebox{0.999\linewidth}{!}{
\begin{tabular}{lcccccccccc}
\toprule
\multirow{3}{*}{\textbf{Policy}} & \multicolumn{2}{c}{\textbf{Scenario 1}} & \multicolumn{2}{c}{\textbf{Scenario 2}} & \multicolumn{2}{c}{\textbf{Scenario 3}} & \multicolumn{2}{c}{\textbf{Scenario 4}} & \multicolumn{2}{c}{\textbf{Avg. Scores (\%)}} \\
\cmidrule(lr){2-3} \cmidrule(lr){4-5} \cmidrule(lr){6-7} \cmidrule(lr){8-9} \cmidrule(lr){10-11} 
& \multicolumn{2}{c}{(Back \twemoji{banana}, Rotate \twemoji{carrot})} & \multicolumn{2}{c}{(Plate \twemoji{ear of corn}, Bowl \twemoji{eggplant})} & \multicolumn{2}{c}{(Poke \twemoji{sponge}, Nudge \twemoji{automobile})} & \multicolumn{2}{c}{(Drop \twemoji{toothbrush}, Forward \twemoji{cucumber})} & \multirow{2}{*}{Seen} & \multirow{2}{*}{Transfer} \\
\cmidrule(lr){2-3} \cmidrule(lr){4-5} \cmidrule(lr){6-7} \cmidrule(lr){8-9} 
& Seen & Transfer & Seen & Transfer & Seen & Transfer & Seen & Transfer & & \\
\midrule
\textbf{\otter} & 97.2 & {33.3} \textcolor{red}{\ding{56}} & 91.7 & 25.0 \textcolor{red}{\ding{56}} & 94.4 & 30.6 \textcolor{red}{\ding{56}} &91.7  & 33.3  \textcolor{red}{\ding{56}} & 93.8 & 30.6 \textcolor{red}{\ding{56}}\\
\pizerofivebold & 94.4 &{41.7} \textcolor{red}{\ding{56}} & 97.2 & 27.8 \textcolor{red}{\ding{56}}  & 97.2 & 44.5 \textcolor{red}{\ding{56}} & 94.4 & 38.9 \textcolor{red}{\ding{56}}& 95.8 & 38.2 \textcolor{red}{\ding{56}}\\
\midrule
\textbf{\ourvla} & 94.4 & {91.7} \textcolor{green}{\ding{52}} & 94.4 & 94.4 \textcolor{green}{\ding{52}} & 97.2 & 91.7 \textcolor{green}{\ding{52}}& 94.4 & 88.9 \textcolor{green}{\ding{52}} & 95.1 & 91.7 \textcolor{green}{\ding{52}}\\
\bottomrule
\end{tabular}
}
\label{tab:real_world_results}
\end{table*}

\begin{figure*}[t]
    \centering
    \begin{subfigure}[b]{0.397\textwidth}
        \centering
        \includegraphics[width=0.98\textwidth]{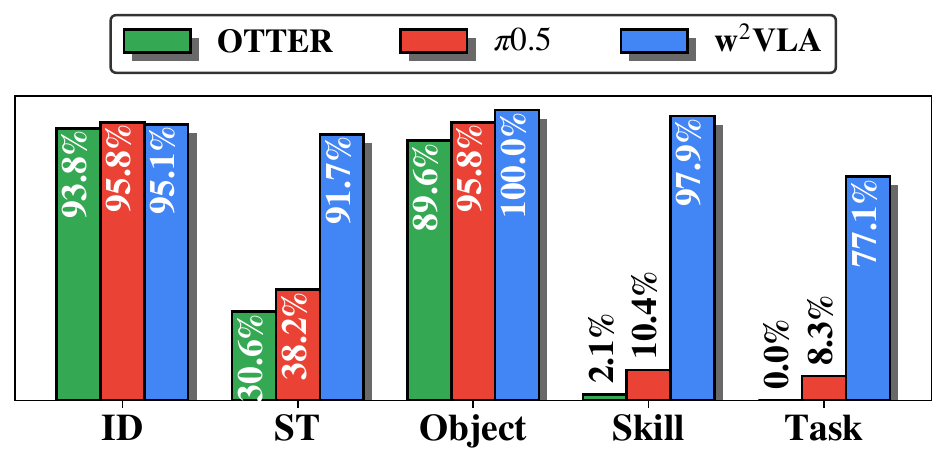}
        \caption{Average VLA performance.}
        \label{fig:left_image}
    \end{subfigure}
    \hfill 
    \begin{subfigure}[b]{0.593\textwidth}
        \centering
        \includegraphics[width=0.98\textwidth]{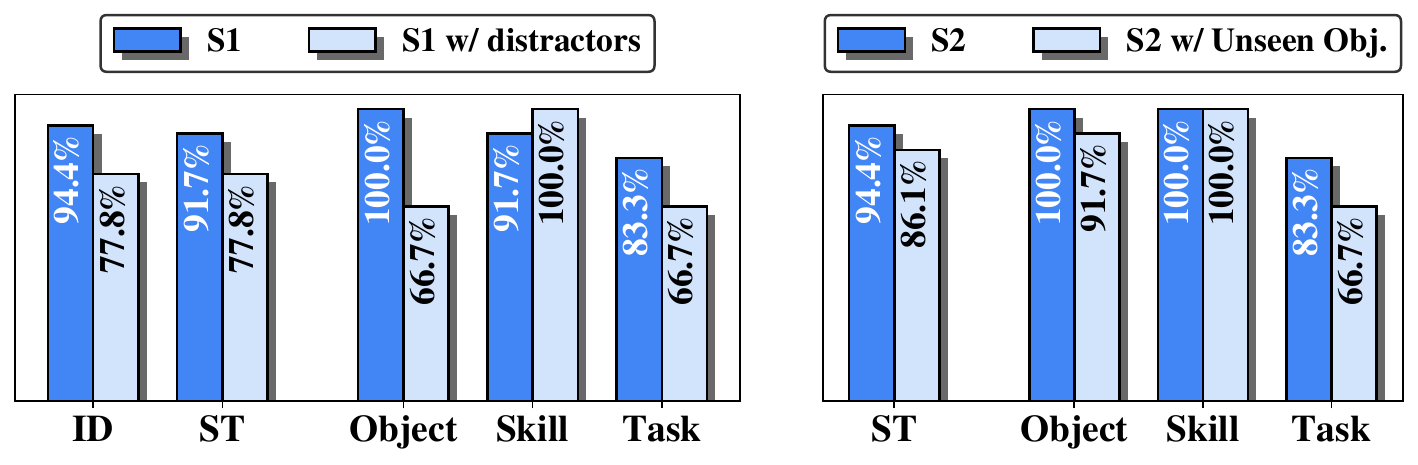}
        \caption{\ourvla robustness experiment results.}
        \label{fig:right_image}
    \end{subfigure}
    \vspace{-15pt}
    \caption{Breakdown of average performance scores: (a) from \otter, \pizerofive, and \ourvla in-domain (ID), \ie on seen \textit{(skill, object)} pairs and on skill transfer (ST), from the experiments of Table~\ref{tab:real_world_results}; (b) from the robustness evaluation of \ourvla, presented in Sec.~\ref{ssec:robustness}. The metrics ``object'', ``skill'', and ``task'' correspond to the success of the models in selecting the correct object, executing the correct skill, and fully completing the task, respectively.}
    \label{fig:avg_barplots}
    \vspace{-20pt}
\end{figure*}

\section{Experiments}
\label{sec:experiments}
We compare \ourvla against \otter and \pizerofive using a real-world SO-101 robot and the LeRobot framework~\cite{cadene2024lerobot}. 
Sec.~\ref{ssec:training_data} outlines the skill transfer scenarios setting. 
Sec.~\ref{ssec:main_experiments} evaluates all three VLAs on both in-domain (\ie seen \textit{(skill, object)} pairs) and skill transfer (\ie applying skills to other objects in the scene). 
Sec.~\ref{ssec:robustness} assesses \ourvla's robustness in both settings, given distractor or completely unseen target objects.
Sec.~\ref{ssec:ablation} ablates \ourvla and finally, Sec.~\ref{ssec:sim} evaluates our model's multi-task capabilities in simulation.

\noindent\textbf{Implementation}: We use DINOv3 ViT-B/16~\cite{siméoni2025dinov3} as the VFM and MetaCLIP2 ViT-L/16 as the two-tower VLM, both frozen, yielding $\sim$55M trainable parameters in \ourvla.
All policies observe a history of $T=8$ steps and execute 10 actions per inference.
We train \ourvla and \otter identically for 15k steps with batch size 32 on a single RTX 4090, using AdamW with a peak learning rate of $1\times10^{-4}$, cosine-annealed to $1\times10^{-5}$ after 200 warmup steps.
For a fair comparison, we replace \otter's default CLIP backbone with the same MetaCLIP2 encoder ($\sim$67M trainable parameters), and both models extract text-aligned patch features identically, following ClearCLIP~\cite{lan2024clearclip}.
For \pizerofive we adopt knowledge insulation~\cite{driess2026knowledge}, freezing the 3B PaliGemma backbone and fine-tuning only the flow-matching action expert and its projection layers ($\sim$700M trainable parameters), for 40k steps with batch size 8, a peak learning rate of $1.5\times10^{-5}$, and a chunk size of 50.
Full end-to-end fine-tuning overfits severely in our low-data regime.

\subsection{Skill Transfer Evaluation}
\label{ssec:training_data}
We use four unique scenarios for collecting data and evaluating the skill transfer capabilities of the studied VLAs.
Visualizations of the scenarios are provided in Fig.~\ref{fig:object_skill_pairs}.
Each scenario consists of two \textit{(skill, object)} pairs, in which 16 expert demonstrations are collected for each, with the two objects in varied poses.

To gain deeper insight into the perception and skill transfer capabilities of each VLA, we designed a granular scoring system. 
In each deployment, a trained policy can score up to three points: one point for interacting with the correct object (evaluating the successful utilization of the declarative knowledge encoded in the VLM), one point for executing the correct behavior (evaluating the policy's procedural knowledge), and a final point for successfully completing the task.

\subsection{Skill Transfer via Compositional Generalization}
\label{ssec:main_experiments}
We evaluate the three VLAs on the seen and unseen \textit{(skill, object)} pairs of each scene, six times for each combination, varying the initial object poses for every rollout. 
We report results in Table~\ref{tab:real_world_results}, which summarizes the success score of each model in each of the four scenarios.
First, we observe that all models manage to showcase comparable, strong performance for all the seen \textit{(skill, object)} pairs, which provides evidence that the compositional design of \ourvla can keep up with state-of-the-art VLAs. 
Second, both \otter and \pizerofive fail to match \ourvla's skill transfer ability, which demonstrated comparable performance to the in-domain cases (\ie 91.7\%), scoring on average 30.6\% and 38.2\%, respectively. 
The analysis of Fig.~\ref{fig:left_image}, sheds more light to this performance gap. 
Even though all three VLAs successfully interact with the correct object, only \ourvla generates the instructed skill.
On the other hand, in the vast majority of skill transfer cases, both \otter and \pizerofive recede to imitating the skill that was paired with the target object during training  (\eg for the \textit{unseen} instruction \textit{(rotate by 90$^o$, banana)} \otter and \pizerofive would execute the \textit{seen} \textit{(place back by 5cm, banana)}).
These results substantiate our claim that modern VLAs are susceptible to learning spurious correlations between the \textit{declarative} and the \textit{procedural}, an issue \ourvla addresses without model and dataset scaling.

\subsection{\texorpdfstring{w$^2$VLA}{w²VLA} Robustness}
\label{ssec:robustness}
We test the performance limits of \ourvla in two out-of-distribution cases by re-deploying our trained model.

\noindent\textbf{Robustness to distractors}: We repeat Scenario 1 with 3-5 visually diverse distractor objects added to the scene.
\ourvla retains competitive performance, dropping by 16.6\% in-domain and 13.9\% under skill transfer.
As shown in Fig.~\ref{fig:right_image}, this is driven by a lower interaction success rate with the target object (100\% to 66.7\%), which propagates to task completion.
Failures were imprecise physical interactions (\eg grasping mid-air) rather than wrong object selection, and the VLM heatmaps remain sharp despite the clutter, consistent with the known sensitivity of IL policies to visual perturbations~\cite{10611331,Hansen2022pre,burns2024what,houlsby2019parameter,spawnet,tsagkas2025attentivefeatureaggregationor}.
Crucially, the instructed skill is executed in every rollout, in both settings.

\begin{figure}[t]
    \centering
    \includegraphics[width=0.98\linewidth]{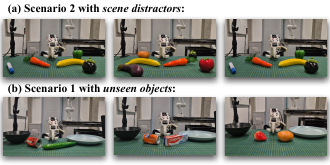}
    \vspace{-5pt}
    \caption{\textbf{Robustness evaluation.}
    }
    \label{fig:vertical_layout}
    \vspace{-24pt}
\end{figure}

\begin{table*}[t]
\centering
\vspace{5pt}
\caption{\textbf{LIBERO-Plus performance}. A single policy trained on all 40 LIBERO tasks, evaluated on canonical states (\textit{Original}) and along seven perturbation axes, with the \textit{drop from Original} shaded below each row. 
}

\label{tab:libero_plus}
\vspace{-4pt}
\small\setlength{\tabcolsep}{4pt}
\renewcommand{\arraystretch}{0.85}
\begin{tabular*}{0.95\textwidth}{@{\extracolsep{\fill}}lrcccccccc||r@{}}
\toprule
Method & Params & Original & Camera & Robot & Lang. & Light & BG & Noise & Layout & Total \\
\midrule
\midrule
\pizerofivebold (from~\cite{han2026geometricactionmodelrobot}) & 3.3B & 96.9 & \textbf{72.0} & \textbf{76.6} & 86.5 & \textbf{96.1} & \textbf{95.2} & 86.7 & \textbf{86.0} & \textbf{84.6} \\
& & &\dcell{25} &\dcell{20} &\dcell{10} &\dcell{1} &\dcell{2} &\dcell{10} &\dcell{11} &\dcell{12} \\
\midrule
\otter (CLIP) & 45M & 71.2 & 43.2 & 47.6 & 15.0 & 66.5 & 65.7 & 63.8 & 53.8 & 50.8 \\
& & &\dcell{28.0} &\dcell{23.5} &\dcell{56.1} &\dcell{4.6} &\dcell{5.4} &\dcell{7.4} &\dcell{17.4} &\dcell{20.4} \\
\midrule
\otter (MetaCLIP2) & 67M & 77.2 & 49.1 & 47.6 & 20.6 & 70.8 & 68.6 & 60.4 & 52.3 & 51.4 \\
& & &\dcell{28.1} &\dcell{29.6} &\dcell{56.6} &\dcell{6.4} &\dcell{8.6} &\dcell{16.8} &\dcell{24.9} &\dcell{25.8} \\
\midrule
\ourvla (privileged) & {56M} & 88.6 & 9.8 & 59.9 & \textbf{88.6} & 87.0 & 87.5 & \textbf{88.1} & 67.3 & 69.7 \\
& & &\dcell{78.8} &\dcell{28.7} &\dcell{0.0} &\dcell{1.6} &\dcell{1.1} &\dcell{0.5} &\dcell{21.3} &\dcell{18.8} \\
\bottomrule
\end{tabular*}
\vspace{-14pt}
\end{table*}
\begin{table}[t]
    \centering
    \caption{\textbf{\ourvla ablations} (Scenario 2): visual modulation (VM),
    patch masking, and module order, on seen \textit{(skill, object)} pairs and
    skill transfer (ST).}
    \setlength{\tabcolsep}{4pt}
    \renewcommand{\arraystretch}{0.9}
    \begin{tabular*}{\columnwidth}{@{\extracolsep{\fill}}lcc@{}}
        \toprule
        Configuration & Seen (\%) & ST (\%) \\
        \midrule
        \ourvla \textit{where}$\rightarrow$\textit{what}, no VM      & 66.7 & 52.8 \\
        \ourvla \textit{where}$\rightarrow$\textit{what}, VM         & 100  & 58.3 \\
        \ourvla \textit{what}$\rightarrow$\textit{where}, VM+Mask    & 100  & 55.6 \\
        \textbf{\ourvla \textit{where}$\rightarrow$\textit{what}, VM+Mask (full)} & \textbf{100} & \textbf{94.4} \\
        \midrule
        \otter (no mask, Scenario 2)      & 91.7 & 25.0 \\
        \pizerofive (no mask, Scenario 2) & 97.2 & 27.8 \\
        \bottomrule
    \end{tabular*}
    \label{tab:ablation}
    \vspace{-6mm}
\end{table}

\noindent\textbf{Skill transfer to novel objects}: We repeat Scenario 2, replacing the corn and eggplant with semantically and visually distinct unseen objects (\eg a soda can and toothpaste).
\ourvla loses only 8.3\%, selecting the correct object in the vast majority of rollouts and executing the instructed skill in all of them, which we attribute to the preserved declarative priors of the frozen VLM.
Task completion drops by 16.7\%, in every case because the geometric gap between seen and unseen objects calls for a trajectory adaptation the policy was never given (\eg picking a mug versus a bottle).

\begin{figure}[t]
    \centering
    \vspace{-3pt}
    \includegraphics[width=0.96\linewidth]{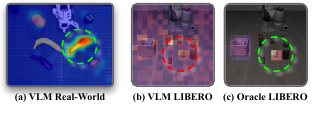}
    \vspace{-10pt}
    \caption{\textit{Where} heatmaps from VLM and Oracle.}
    \label{fig:libero_heatmaps}
    \vspace{-20pt}
\end{figure}

\noindent\textbf{VLM Localization Accuracy}: \ourvla's \textit{where} module relies on the frozen VLM's text-aligned heatmap to localize the target object, which is highly precise in the real world (Fig.~\ref{fig:libero_heatmaps}a). 
However, it fails when an asset does not visually match its name. 
In LIBERO, several meshes do not depict the object their instruction names (\eg the ``cream cheese'' is a featureless blue box), leaving nothing to match and erasing the \textit{where} signal (Fig.~\ref{fig:libero_heatmaps}b). 
Because the VLM is frozen and the design is compositional, \ourvla \textit{cannot} compensate by overfitting to such text-image pairs, a failure we attribute to LIBERO's asset fidelity rather than the backbone.

\subsection{\texorpdfstring{w$^2$VLA}{w²VLA} Ablation}
\label{ssec:ablation}
We ablate \ourvla on Scenario 2 (Table~\ref{tab:ablation}).
Relying on heatmaps alone (no Visual Modulation) still qualitatively enables skill transfer, indicating that the spatial and skill conditioning modules undertake a significant share of the action prediction process, but manipulation becomes imprecise and both success rates drop.
Adding the VFM signal restores in-domain performance, yet transfer remains unreliable: without direct supervision the policy infers the skill from appearance, reintroducing the spurious correlations we aim to break.
Random patch masking removes this shortcut.

Neither ingredient is sufficient in isolation.
Without masking, \ourvla still reaches 58.3\% transfer on this scenario, against 25.0\% for \otter and 27.8\% for \pizerofive (Table~\ref{tab:real_world_results}), which are trained without masking on the same data.
Conversely, the \textit{what}$\rightarrow$\textit{where} variant is trained \textit{with} masking and still collapses to 55.6\%, so masking alone does not produce transfer either.
The two are complementary: masking suppresses the appearance shortcut, and the decoupled \textit{where}$\rightarrow$\textit{what} ordering supplies the signal the policy falls back on once that shortcut is gone.

Module order matters equally. 
Applying \textit{where} first grounds the region of interaction before \textit{what} specifies the behavior. 
Reversing the two collapses transfer to 55.6\%, as the inherently noisy dense visual signal can trigger the wrong behavior before the target is localized. 
This mirrors human sensorimotor coordination, where locating an object precedes planning a targeted interaction.

\subsection{Compositionality and Multi-skill Learning}
\label{ssec:sim}
The experiments above operate in a deliberately small-data regime, where compositional priors matter most. 
We now verify that the proposed decomposition does not sacrifice the representation power required for more complex multi-task learning scenarios and dataset scaling, that generic VLAs are expected to tackle. 
We train a single \ourvla on all four LIBERO suites~\cite{liu2023libero} (40 tasks) and evaluate the same checkpoint on LIBERO-Plus~\cite{fei25libero-plus}, which perturbs those suites along seven axes, without any retraining.

All 40 LIBERO instructions are expressed as $(l_{skill}, l_{obj})$ commands, with no task excluded or per-task tuning.
The split is mechanical: verbs are routed to \textit{what}, while entities to be localized are routed to \textit{where}, keeping the two slots disjoint so that the \textit{what} module cannot recover localization by memorizing skill-object pairings.
This covers a standard benchmark's full task distribution, including the long-horizon suite, showcasing that the proposed command interface is adequately expressive to drive a low-level policy along complex tasks. 
Furthermore, in order to isolate the impact of the restructured information flow of our policy and mitigate failures caused by the VLM localization issue of LIBERO (Sec.~\ref{ssec:robustness}), in this experiment we adopt privileged information of object footprints from LIBERO's segmentation buffer, to restore spatial localization of the affected objects. 

In this setting, a single \ourvla trained on all 40 LIBERO tasks reaches $88.6\%$ (Table~\ref{tab:libero_plus}), striking a balance between \otter's ($77.2\%$) ingenuous adoption of a frozen VLM backbone, and \pizerofive ($96.9\%$) which carries $3.3$B parameters and large-scale robot pre-training, against $56$M trained from scratch. 
Under the seven LIBERO-Plus perturbations \ourvla retains $69.7\%$, degrading by $18.8$ points against $25.8$ for \otter and $12.0$ for \pizerofive. 
This shows that compositional structure maintains the strong multi-task scaling and perturbation robustness capabilities, expected of a generic VLA.


\ourvla shows no drop under language perturbations, as the high-level planner maps rephrased instructions to the same atomic (skill, object) commands before they reach the policy. 
It also takes negligible drops ($<2\%$) under background, lighting, and sensor noise perturbations, which we attribute to the \textit{where} and \textit{what} conditioning modules undertaking a notable portion of the action prediction process, making it more resilient to visual perturbations.
Camera shift, however, is catastrophic ($\downarrow 78.8$), since the \textit{where} signal is defined on the image patch grid, the policy learns a mapping from heatmap location to end-effector motion that holds only under the training camera extrinsics. 
Grounding \textit{where} in global 3D coordinates (\eg lifting the heatmaps into the robot base frame using depth and camera calibration) would make this signal viewpoint-invariant, while pre-training the action expert on OXE with the VLM encoders kept frozen could supply the viewpoint and embodiment diversity that benefits \pizerofive, combining the best of both worlds.

\section{Discussion}
\label{sec:discussion}
We examined the information flow of popular VLAs and showed that, even in minimal scenarios with two \textit{(skill, object)} pairs, they fail to transfer a learned skill to another object.
Although these models select the correct object, they execute the behavior paired with it during training, revealing an entanglement of declarative and procedural knowledge.
\ourvla resolves this by sequentially conditioning action prediction on decoupled representations of the \textit{where} of the target object and the \textit{what} of the desired behavior, achieving reliable zero-shot skill transfer while matching in-domain performance.
The same interface covers all 40 LIBERO tasks, including the long-horizon suite, and an off-the-shelf VLM decomposes abstract instructions into its commands zero-shot.
This makes \ourvla a natural low-level policy for hierarchical frameworks~\cite{shi2025hi,torne2026memmultiscaleembodiedmemory,chen2026steerable}.
The main open challenge is transfer between objects with substantially different geometry, where the trajectory itself must adapt (\eg picking up a mug after learning to pick up a bottle).
A promising extension is a \textit{how} module that conditions the policy on fine-grained geometric features, such as grasp poses from AnyGrasp~\cite{fang2023anygrasp}, or adapts learned primitives to new geometries from a few additional demonstrations.
Ultimately, we hope these findings encourage the community to look beyond scaling models and datasets, towards the structural priors needed for generalist robot policies.




\section*{Acknowledgement}
This work was supported by the United Kingdom Research and Innovation (grant EP/S023208/1), EPSRC Centre for Doctoral Training in Robotics and Autonomous Systems (RAS). 
Part of this work was conducted while Nikolaos Tsagkas was conducting an internship at the Samsung AI Center in Cambridge, UK.\newline

\bibliographystyle{IEEEtran} 
\bibliography{bibliography}

\setcounter{section}{0}
\renewcommand{\thesection}{A\arabic{section}}
\setcounter{figure}{0}
\setcounter{table}{0}
\setcounter{equation}{0}
\renewcommand{\thefigure}{A\arabic{figure}}
\renewcommand{\thetable}{A\arabic{table}}
\renewcommand{\theequation}{A\arabic{equation}}
\section*{APPENDIX}

\section{Related Work}
\label{sec:related_work}

\noindent\textbf{Vision-Language-Action Models}.
Inspired by the broad success that LLMs and VLMs have achieved by scaling both model size and training data, the robotics community has increasingly adopted this paradigm in pursuit of a robot foundation model.

Currently, the dominant approach for training VLAs involves fine-tuning powerful, pre-trained LLMs and VLMs. 
Notable examples in this category include: 
(1) OpenVLA~\cite{pmlr-v270-kim25c}, which maps visual and language embeddings to tokens that are processed by a Llama 2 \cite{touvron2023llama} backbone, fine-tuned to output action deltas. 
(2) \pizero~\cite{black2024pi_0}, which utilizes a pre-trained VLM backbone (PaLiGemma) incorporating SigLIP~\cite{zhai2023sigmoid} and Gemma encoders to generate action deltas via a flow-matching action head. 
(3) \pizerofive~\cite{black2025pi05}, which extends \pizero~\cite{black2024pi_0} by co-training on heterogeneous datasets. 
(4) VLA-0~\cite{goyal2025vla0buildingstateoftheartvlas}, which represents actions as text to ensure the vocabulary of the backbone VLM remains undiluted.

Recently, \otter~\cite{huang2025otter} challenged this paradigm by employing a two-tower VLM backbone (\ie CLIP~\cite{radford2021learning} and ClearCLIP~\cite{lan2024clearclip}) and keeping it entirely frozen during policy training, preserving the encoded declarative knowledge. 
\otter extracts text-aware visual features and concatenates them with proprioceptive and language features before passing them to a causal transformer. 
We argue, however, that this naive feature concatenation can result in the model learning spurious correlations between visual observations and the skill being executed.

We benchmark our approach against \otter and \pizerofive, selecting them as representative state-of-the-art VLA baselines that encapsulate two distinct paradigms. 
Specifically, \otter represents architectures that freeze the VLM backbone to leverage pre-trained declarative semantics, whereas \pizerofive exemplifies the end-to-end training paradigm driven by large-scale parameter and dataset scaling.

Our approach draws inspiration from \otter, as we also utilize a frozen, two-tower VLM as the backbone for \ourvla. 
However, we propose a novel compositional architecture for conditioning the policy on the object of interest and the desired behavior. 
By doing so, we prevent these spurious correlations and successfully unlock skill transfer capabilities in VLAs.

\noindent\textbf{Robot Learning via Minimal Visual Cues}.
\label{ssec:minimal_vis_cues}
In contrast to recent trends in robot learning, a promising paradigm avoids feeding multiple, rich tokens extracted from large pre-trained foundation models directly into policies. 
Instead, the intuition is  that visual inputs must first be abstracted into a form that is more usable and interpretable for robot policies and planning algorithms.

In the context of imitation learning, CLIPort~\cite{shridhar2021cliport} formulated manipulation as discrete, two-step spatial primitives conditioned by predicted affordance maps. 
Similarly, KITE~\cite{sundaresan2023kite} relied on extracted keypoints to execute learned, keypoint-conditioned skills. 
More recently, PEEK~\cite{zhang2026peek} was designed to extract essential keypoints and 2D gripper paths, explicitly overlaying these cues onto the robot's visual input.

Within planning-based approaches, a series of works have fused VFM and VLM features into spatial representations to reason about areas  of interest in a scene and plan manipulation trajectories.
VoxPoser~\cite{huang2023voxposer} utilized VLMs to compose 3D value maps, grounding language-aligned knowledge into the agent's observation space, while VL-Fields~\cite{tsagkas2023vlfields} fused VLM features to segment in an open-vocabulary manner objects of interest in the scene. 
Furthermore, F3RM~\cite{shen2023F3RM}, C2G~\cite{tsagkas2024click} fused VFM features into implicit spatial representations to localize interaction areas and optimize end-effector poses for object manipulation.
Concurently with our work, GROW$^2$~\cite{deng2026grow2groundingrobottool} extended this idea to tool use, employing a VLM to reason about which object should serve as the tool and which of its parts are task-relevant, before grounding where to interact by lifting those parts into 3D regions with VFMs.

However, by explicitly decoupling perception and planning into multi-stage pipelines or discrete action spaces, these methods sacrifice the high-dimensional, continuous control capabilities inherent to modern VLAs.
Drawing inspiration from these methods, our work conditions the policy on low-level visual information (\ie localization heatmaps for the object of interest). 
This abstracts the raw visual signal, guiding the model's attention strictly to the relevant areas.

\noindent\textbf{Skill Transfer}. In the context of VLAs, skill transfer is often indirectly evaluated through a model's ability to generalize to unseen instructions. 
For instance, RT-1~\cite{brohan2023rt1roboticstransformerrealworld} evaluated zero-shot generalization across novel tasks, environments, and objects, demonstrating notable robustness. 
However, this generalization was explicitly achieved by scaling the training data to a massive volume of roughly 130,000 real-world robot trajectories covering over 700 distinct tasks, collected over 17 months. 

Similarly, {RT-2}~\cite{zitkovich2023rt} almost doubled generalization performance to unseen objects. 
This was achieved by co-fine-tuning large vision-language models on Internet-scale web data alongside robot data and expressing robot actions as text tokens, thereby inheriting broad semantic reasoning but relying heavily on massive model and data scaling. 
{OCTO}~\cite{octo_2023} tackled cross-embodiment generalization by pretraining a diffusion policy on an aggregated 800,000 robot trajectories from the Open X-Embodiment dataset. 
More recently, {OpenVLA}~\cite{pmlr-v270-kim25c} achieved state-of-the-art generalist manipulation by training a 7B parameter model on a diverse collection of 970,000 real-world robot demonstrations. 
Similarly, {\pizero} utilized a flow-matching architecture on top of a pre-trained VLM to generate continuous actions. 
Its extension, \pizerofive, introduced co-training on heterogeneous tasks to unlock better generalization, scaling the dataset even more.

As is evident, the current trend in the robot learning community relies on scaling parameters and massive datasets to yield emergent behaviors that afford skill transfer capabilities. 
However, this is computationally expensive and, as our experiments prove, does not natively allow for the zero-shot transfer of procedural mechanics from a single \textit{(skill, object)} pair in low-data regimes. 
Recently, {OTTER} demonstrated that VLAs can diverge from the massive-scale finetuning paradigm by freezing the two-tower VLM backbone to effectively utilize pre-trained declarative knowledge.
As an emergent property, they reported successful execution of unseen instructions (\eg executing a seen skill on an unseen object). 
Nevertheless, we argue that this is largely the result of visual interpolation. 
For example, their ``poke'' and ``pour'' primitives were trained across multiple distinct object and color combinations, enabling the model to interpolate a generalized physical behavior onto a new semantic target within a dense training manifold. 
Our experiments showcase that such an architecture, which concatenates features into a unified sequence, is still susceptible to learning spurious correlations between objects and demonstrated skills when restricted to a low-data envelope. 
In contrast, \ourvla advocates for the incorporation of internal compositional structure within VLAs, sequentially decoupling the spatial \textit{where} from the procedural \textit{what}, rather than relying on massive, opaque, end-to-end models and dataset scaling alone.

\section{Skill Transfer Scenarios}
\label{app:real_world}
\label{app:skill_transfer_scenarios}
In this section, we provide additional details about the four skill transfer scenarios we designed capabilities of VLAs. 
Each scenario consists of two unique \textit{(skill, object)} pairs. 

\begin{figure}[h]
    \centering
    \includegraphics[width=0.9\linewidth]{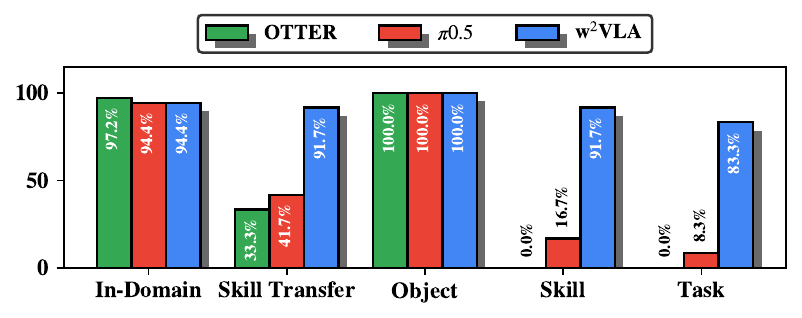}
    \caption{\textbf{Scenario 1 average results}: Average scores for the \otter, \pizero, and \ourvla for the seen \textit{(skill, object)} pairs and the skill transfer case. For the latter, we also provide average scores for: choosing the correct object to interact with (``Object''), imitating the correct behavior (``Skill''), and finally, managing to complete the task (``Task'')}
    \vspace{-10pt}
    \label{fig:scenario_1_barplot}
\end{figure}

\noindent\textbf{Scenario 1 --- \textit{(place back 5cm, banana \twemoji{banana})} \textit{\&} \textit{(rotate by 90$^o$, carrot \twemoji{carrot}})}: The banana and the carrot lie in front of the robot base.
We collect 16 demonstrations in which the robot picks, lifts, and rotates the carrot by ${90^{\circ}}$ counterclockwise, before placing it back down in the same location.
An additional 16 demonstrations are collected where we pick up and place back the banana by roughly 5cm. 
We consider task completion successful if the robot managed to pick up the object and, depending on the given instruction, rotated the it by roughly 90$^o$ or placed it back by roughly 5cm.
In Fig.~\ref{fig:scenario_1_barplot}, we present the performance for the three VLAs for the seen \textit{(skill, object)} pairs and for skill transfer.

\begin{figure}[h]
    \centering
    \includegraphics[width=0.9\linewidth]{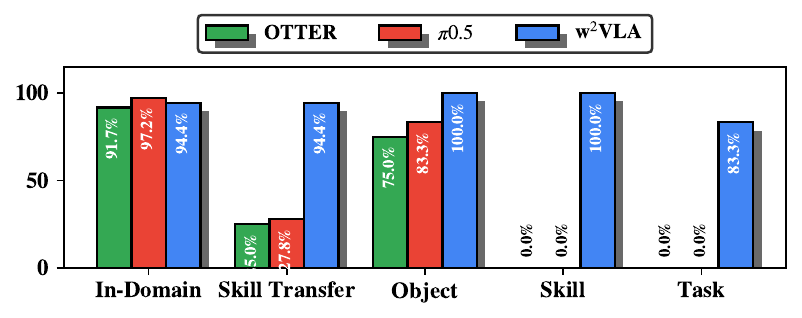}
    \vspace{-10pt}
    \caption{\textbf{Scenario 2 average results}: Average scores for the \otter, \pizero, and \ourvla for the seen \textit{(skill, object)} pairs and the skill transfer case. For the latter, we also provide average scores for: choosing the correct object to interact with (``Object''), imitating the correct behavior (``Skill''), and finally, managing to complete the task (``Task'')}
    \vspace{-10pt}
    \label{fig:scenario_2_barplot}
\end{figure}

\noindent\textbf{Scenario 2 --- \textit{(place on plate, corn \twemoji{ear of corn})} \textit{\&} \textit{(place on bowl, eggplant \twemoji{eggplant})}}: The corn and the eggplant are placed in front of the robot base.
Also, a plate and a bowl are positioned on the left and right of the robot base, respectively, at a distance of approximately 10cm.
We collect 16 demonstrations in which the eggplant is picked and placed into the black bowl, and 16 more for picking and placing the corn on to the light-blue plate. 
We consider task completion successful if the robot managed to pick up the object and, depending on the given instruction, placed it in the bowl or plate.
In Fig.~\ref{fig:scenario_2_barplot}, we present the performance for the three VLAs for the seen \textit{(skill, object)} pairs and for skill transfer.

\begin{figure}[h]
    \centering
    \includegraphics[width=0.9\linewidth]{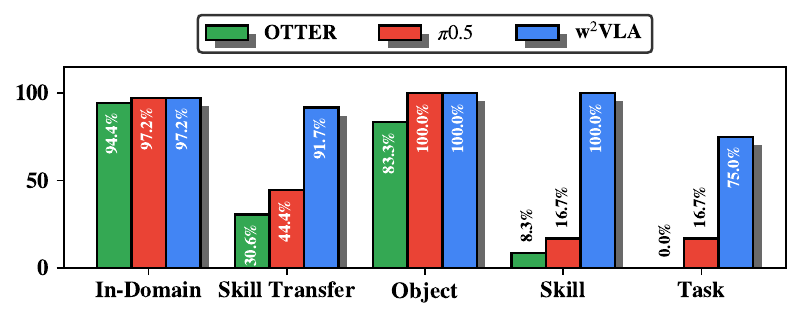}
    \vspace{-10pt}
    \caption{\textbf{Scenario 3 average results}: Average scores for the \otter, \pizero, and \ourvla for the seen \textit{(skill, object)} pairs and the skill transfer case. For the latter, we also provide average scores for: choosing the correct object to interact with (``Object''), imitating the correct behavior (``Skill''), and finally, managing to complete the task (``Task'')}
    \vspace{-10pt}
    \label{fig:scenario_3_barplot}
\end{figure}

\noindent\textbf{Scenario 3 --- \textit{(poke, sponge \twemoji{sponge})} \textit{\&} \textit{(nudge, toy car \twemoji{automobile})}}: The sponge and the toy car are placed in front of the robot base.
We collect 16 demonstrations in which the robot approaches the sponge from above with the gripper fingers closed, descents to poke it, and then ascents again. 
Another 16 are collected in which the gripper, with the fingers closed again, approaches the toy car from the back and swings the entire end-effector to nudge the object.
We consider task completion successful if the robot managed to, depending on the given instruction, swing the end-effector from the back of the object, making it flip forward, or coming down vertically on the top of the object.
In Fig.~\ref{fig:scenario_3_barplot}, we present the performance for the three VLAs for the seen \textit{(skill, object)} pairs and for skill transfer.

\begin{figure}[h]
    \centering
    \includegraphics[width=0.9\linewidth]{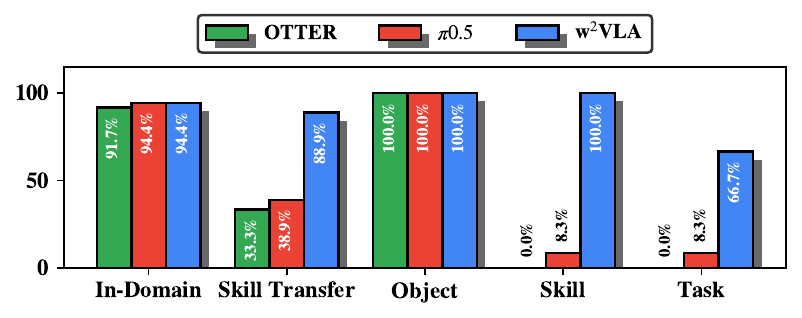}
    \vspace{-10pt}
    \caption{\textbf{Scenario 4 average results}: Average scores for the \otter, \pizero, and \ourvla for the seen \textit{(skill, object)} pairs and the skill transfer case. For the latter, we also provide average scores for: choosing the correct object to interact with (``Object''), imitating the correct behavior (``Skill''), and finally, managing to complete the task (``Task'')}
    \vspace{-10pt}
    \label{fig:scenario_4_barplot}
\end{figure}

\noindent\textbf{Scenario 4 --- \textit{(drop on plate, toothpaste \twemoji{toothbrush})} \textit{\&} \textit{(place forward, cucumber \twemoji{cucumber})}}: The two objects are positioned in front of the robot base, and a paper plate is positioned approximately 10cm on the right of the robot. 
We collect 16 demonstrations in which the robot picks up the toothpaste box, brings it on top of the plate and drops it. 
Another 16 are collected in which the robot picks up the cucumber and places it 5cm forward.
We consider task completion successful if the robot managed to pick up the object and, depending on the given instruction, drops it on the plate or places it forward by roughly 5cm.
In Fig.~\ref{fig:scenario_4_barplot}, we present the performance for the three VLAs for the seen \textit{(skill, object)} pairs and for skill transfer.

\section{\texorpdfstring{w$^2$VLA}{w²VLA} Robustness Evaluation}
\label{app:robustness}
We provide more information about the \ourvla robustness experiments, described in Sec.~\ref{ssec:robustness} in the main paper.
In Sec.~\ref{app:distractors}, we provide more details about the modification of Scenario 1, where we added additional random distractors to the scene during deployment, which were not present when we collected the training demonstrations. 
In Sec.~\ref{app:distractors}, we  further discuss the modification of Scenario 2, where during deployment we replaced the banana and carrot target objects with unseen, visually different ones.

\subsection{Scenario 1 with Distractor Objects}

\label{app:distractors}
\begin{figure}[h]
    \centering
    \includegraphics[width=1\linewidth]{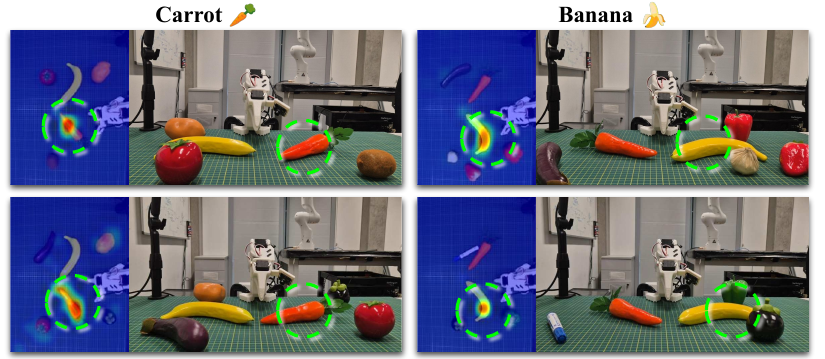}
    \caption{\textbf{Example distractor scenes}: We provide representative scenes from Scenario 1 with random distractor objects. Along with each view, a localization heatmap is provide that showcases the robustness of the backbone VLM in spatially conditioning \ourvla on the area of interaction.}
    \label{fig:distractors}
    \vspace{-15pt}    
\end{figure}

Fig.~\ref{fig:distractors} visualizes four representative scenes of Scenario 1 with random distractor objects, as discussed in Sec.~\ref{ssec:robustness}.
Along with pictures from these cases, we also provide localization heatmaps, which are used for conditioning the \textit{where} module.
As is evident, the features from the deployed VLM (\ie MetaCLIP2), are powerful enough to ignore the distractors and provide feature maps that can condition \ourvla on a specific location for interaction.

\subsection{Scenario 2 with Unseen Objects}
\label{app:unseen}
\begin{figure}[h]
    \centering
    \includegraphics[width=1\linewidth]{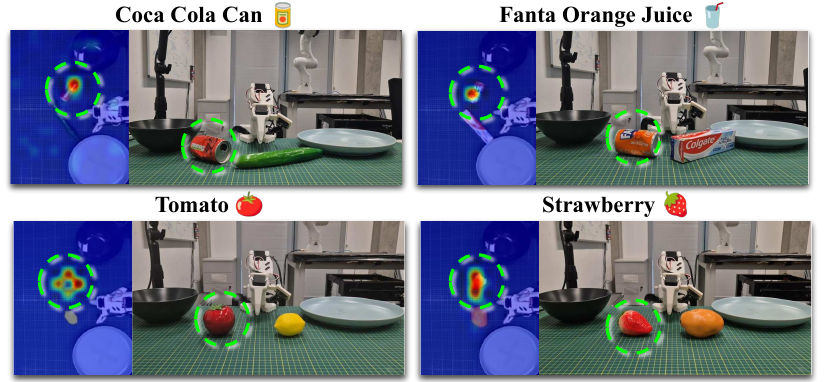}
    \caption{\textbf{Example scenes with unseen objects}: We provide representative scenes from Scenario 2 with unseen target objects. Along with each view, a localization heatmap is provide that showcases the robustness of the backbone VLM in spatially conditioning \ourvla on the area of interaction.}
    \label{fig:unseen}
    \vspace{-15pt}
    
\end{figure}

Fig.~\ref{fig:unseen} visualizes four representative scenes of Scenario 2 with the two original target objects (\ie banana and carrot) with visually diverse ones (\eg a coca cola can), as discussed in Sec.~\ref{ssec:robustness}.
Similarly to Sec.~\ref{app:distractors}, we provide localization heatmaps, which are used for conditioning the \textit{where} module, and showcase that the VLM has no problem localizing the new objects with precision.  
Thus, task completion failure came from the robot manipulating the target object in an inaccurate way, due to geometric discrepancies. 
In Table~\ref{tab:unseen} we summarize the different combinations of unseen \textit{(skill, object)} pairs, providing the complete language instructions. 

\begin{table*}[ht]
    \centering
    \vspace{5pt}
    \caption{List of all \textit{(skill, object)} pairs, with unseen objects from the experiments in Sec.~\ref{app:unseen}, tested during the robustness evaluation of \ourvla. Each column corresponds to a unique rollout.}
    \resizebox{\textwidth}{!}{%
    \begin{tabular}{lcccccccccccc}
        \toprule
        \textbf{Instruction} & \makecell[c]{Place the \\ banana \\ on the plate} & \makecell[c]{Place the \\ carrot \\ in the bowl} & \makecell[c]{Place the \\ lemon \\ on the plate} & \makecell[c]{Place the \\ tomato \\ in the bowl} & \makecell[c]{Place the \\ strawberry \\ in the bowl} & \makecell[c]{Place the \\ potato \\ on the plate} & \makecell[c]{Place the \\ sponge \\ on the plate}  & \makecell[c]{Place the \\ toy car \\ on the plate} & \makecell[c]{Place the \\ cucumber  \\ the plate} & \makecell[c]{Place the \\ toothpaste box  \\ on the plate} & \makecell[c]{Place the \\ Coca Cola can \\  in the bowl} & \makecell[c]{Place the \\ Fanta can \\ in the bowl} \\
        \midrule
        \textbf{Skill}  & Plate & Bowl & Plate & Bowl & Bowl & Plate & Plate & Bowl & Plate & Plate & Bowl & Bowl \\
        \textbf{Object} & Banana & Carrot & Lemon & Tomato & Strawberry & Potato & Sponge & Toy Car & Cucumber & Toothpaste & Coca Cola & Fanta \\
        \bottomrule
    \end{tabular}%
    }
    \vspace{5pt}

    \label{tab:unseen}
    \vspace{-10pt}
\end{table*}

\section{Implementation Details}
\label{app:implementation_details}
\subsection{\texorpdfstring{w$^2$VLA}{w²VLA} and Baselines}

We evaluate our approach against two state-of-the-art baselines representing distinct VLA paradigms: \otter (frozen backbone with explicit feature extraction) and \pizerofive (massive-scale pre-training).
In Fig.~\ref{fig:vla_comparison}, we illustrate how most VLAs utilize the same information differently.

\noindent \textbf{\otter}:
Selected for its conceptual similarities to our method, \otter also preserves pre-trained semantic priors by freezing its two-tower VLM backbone. 
It achieves visual grounding by computing cosine similarities between intermediate visual patch features and language tokens, extracting only the text-aligned visual information. 
These text-aware features are compressed via AFA~\cite{tsagkas2025attentivefeatureaggregationor}, concatenated with the encoded proprioceptive state in a unified sequence, and fed into a causal transformer to predict actions. 
For a rigorously fair comparison, we upgrade \otter's default CLIP backbone to MetaCLIPv2-ViT-L/16, matching the visual encoder used in our model.
Nevertheless, we follow OTTER's method for extracting visual text-aligned features from the visual encoder of a two-tower VLM (inspired by ClearCLIP~\cite{lan2024clearclip}), using the last self-attention block's attention feature map.
Note that this is also how we extract VLM features for \ourvla for creating localization heatmaps to condition the \textit{where} module.
Visualizations like the ones in Fig.~\ref{fig:distractors} and Fig.~\ref{fig:unseen} are evidence that truly these feature maps are well aligned with text, allowing for precise localization of objects, even in cluttered scenes. 
Based on our configuration, \otter consists of 67.11M trainable parameters in total. 

\noindent {$\mathbf{\boldsymbol{\pi}0.5}$}: 
Representing the paradigm of massive, heterogeneously pre-trained VLAs, \pizerofive utilizes a 3B-parameter PaliGemma VLM~\cite{beyer2024paligemmaversatile3bvlm} with joint attention layers between visual, language, and proprioceptive tokens, coupled with a flow-matching action expert. 
As fully fine-tuning such a large model end-to-end in our low-data IL regime risks severe overfitting and catastrophic forgetting, we adopt the \textit{knowledge insulation} strategy~\cite{driess2026knowledge}. 
Specifically, we completely freeze the discrete $3$B-parameter PaliGemma foundation backbone (generally a common finetuning strategy~\cite{shukor2025smolvla,guo2026priorvla,chen2026pointact,bjorck2025gr00t}) to stop gradient flow from the continuous action predictions. 
We restrict fine-tuning exclusively to the action expert and its associated continuous temporal projection layers (\textit{e.g.}, \texttt{action\_in\_proj}, \texttt{time\_mlp\_in}). 
Based on our architectural configuration, this limits the trainable footprint to exactly 693.42M parameters. 
This insulation prevents continuous flow-matching gradients from degrading the VLM's discretely pre-trained semantic representations, preserving its broad generalization capabilities while allowing a fair assessment of its ability to acquire new procedural skills.

\noindent{\textbf{\ourvla}}: 
Our configuration resulted in 55.17M trainable parameters, making it the smaller of the three studied VLAs, even though its size remain comparable to that of \otter.
A breakdown of these parameters is included in Table~\ref{tab:module_params}. 
As mentioned above, \ourvla incorporates the same VLM feature extraction process of ClearCLIP~\cite{lan2024clearclip}.
It should be noted here that \ourvla is readily extensible to incorporate more complex conditioning signals (\eg instead of performing spatial conditioning with 2D heatmaps, one can utilize localization in 3D spatial representations~\cite{tsagkas2024click}) or swap the action head with action flow matching~\cite{bjorck2025gr00t,sochopoulos2025fast,sochopoulos2026optimaltransportqlearningflow,black2025pi05,intelligence2025pi06vlalearnsexperience}.

\begin{table}[t]
    \centering
    \caption{\ourvla number of trainable parameters per module.}
    \label{tab:module_params}
    \begin{tabular*}{\columnwidth}{@{\extracolsep{\fill}}lr@{}}
        \hline
        \textbf{Module} & \textbf{No. Params} \\
        \hline
        Robot State Enc.  & 0.27M \\
        Visual Modulator  & 14.18M \\
        \textit{Where}    & 15.22M \\
        \textit{What}     & 21.52M \\
        Action Head       & 3.99M \\
        \hline
        \textbf{Total}    & \textbf{55.17M} \\
        \hline
    \end{tabular*}
    \vspace{-15pt}
\end{table}

\subsection{Training Details}
Below we report the training recipes for the three VLAs tested.
Given the similarities between \otter and \ourvla, we follow the same recipe for those two models and have a separate discussion for \pizerofive.

\noindent \textbf{\otter} \& \textbf{\ourvla}:
We train both \otter and \ourvla for a total of $15,000$ steps with a batch size of $32$ on an NVIDIA GeForce RTX 4090.
Optimization is conducted via the AdamW~\cite{loshchilov2017decoupled} optimizer with a peak learning rate of $1 \times 10^{-4}$ and a weight decay of $1 \times 10^{-4}$.
The learning rate follows a cosine decay schedule, warming up for 200 steps before stepping down to a minimum of $1 \times 10^{-5}$.
For temporal processing, both architectures ingest an identical observation history length of $8$ steps and feature an action execution horizon of $10$ steps.
Both architectures also route their final representations through a $4$-layer MLP action head configured with a hidden dimension of $512$ to decode action deltas.

\noindent {$\mathbf{\boldsymbol{\pi}0.5}$}: We fine-tune the model for a total of $40,000$ steps with a batch size of $8$ on an NVIDIA GeForce RTX 4090. 
Optimization is performed using the AdamW optimizer with a peak learning rate of $1.5 \times 10^{-5}$, coefficients $\beta = (0.9, 0.95)$, an epsilon value of $\epsilon = 10^{-8}$, and a weight decay of $0.01$. 
Gradients are clipped to a maximum norm threshold of $1.0$. 
The learning rate is regulated via a cosine decay schedule with a warmup phase for the first $1,000$ steps before decaying to a minimum learning rate of $2.5 \times 10^{-6}$ at $40,000$.
For observation and action processing, input camera frames are resized to a target resolution of $224 \times 224$ pixels, and the language tokenizer max sequence length is constrained to $200$ tokens. 
The policy utilizes an action prediction chunk size (horizon) and execution step parameter of $50$.

\subsection{Hardware Information}
\begin{figure}
    \centering
    \includegraphics[width=0.81\linewidth]{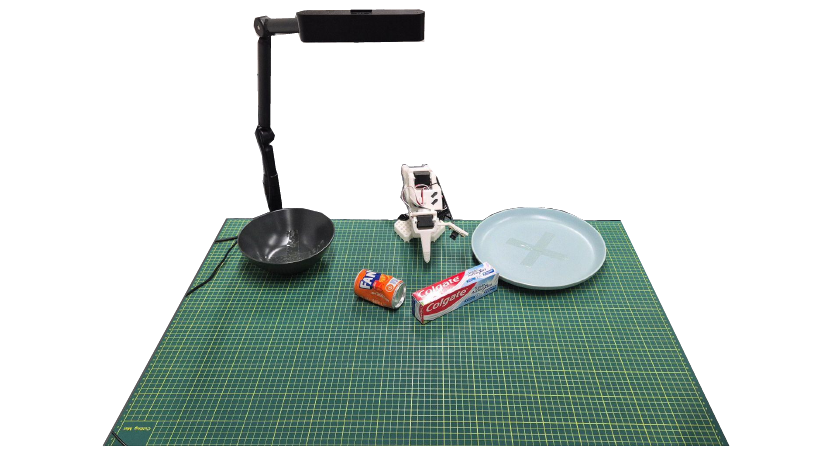}
    \caption{\textbf{Real-world robot setup}: our setups consists of an SO-101 unit and a ZED2i camera, positioned overlooking the scene.}
    \label{fig:hardware_setup}
    \vspace{-15pt}
\end{figure}

We run all our experiments on an SO-101 (leader/follower setup).
We mount a ZED2i stereo camera to overlook the scene in a top-down view.
From this camera we only utilize the input from the left lens and make no use of the depth perception.
The input is center-cropped, with the final resolution being 256x256. 
A visualization of our real-world setup is provided in Fig.~\ref{fig:hardware_setup}

\section{Simulation Evaluation}
\label{sec:appendix-sim}

The real-world experiments of Sec.~\ref{ssec:main_experiments} operate in a
deliberately small data regime, which is where the compositional priors of
\ourvla are expected to matter the most.
Here we examine the complementary setting, in which the number of tasks and the
size of the dataset grow.
We train a single \ourvla policy on all four LIBERO suites~\cite{liu2023libero}
simultaneously (40 tasks) and evaluate
the same checkpoint under systematic visual, physical and linguistic
perturbations on LIBERO-Plus~\cite{fei25libero-plus}.
The purpose of this evaluation is not to make a new architectural claim, but to
verify that the proposed decomposition does not sacrifice the generic
capabilities expected of a VLA, namely multi-task scaling
(Sec.~\ref{sec:appendix-libero}) and robustness to perturbations
(Sec.~\ref{sec:appendix-libero-plus}).

\subsection{LIBERO}
\label{sec:appendix-libero}

\begin{figure*}[ht]
\centering
\begin{subfigure}[b]{0.49\linewidth}
    \centering
    \includegraphics[width=\linewidth]{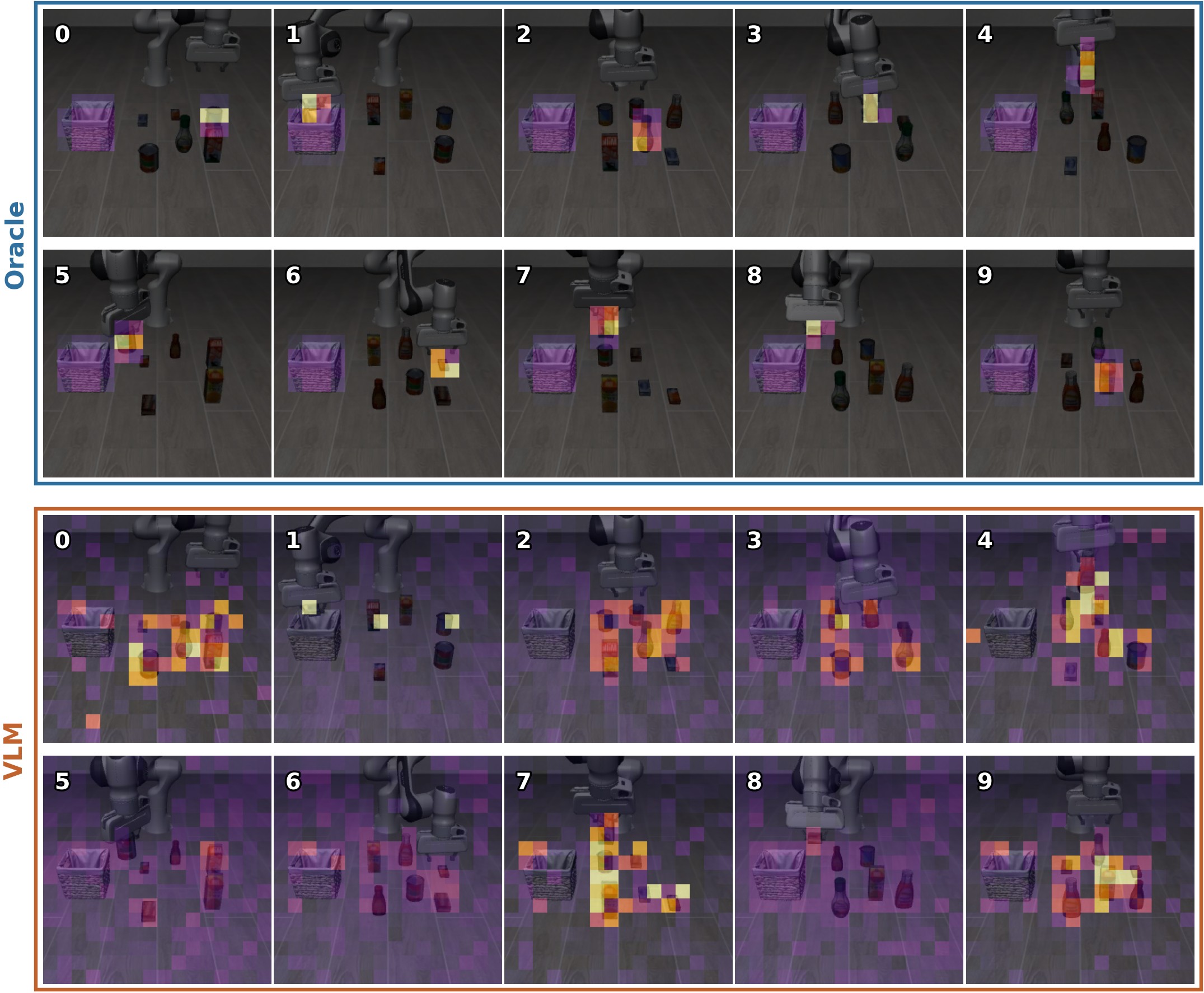}
    \caption{LIBERO-Object}
    \label{fig:appendix-heatmaps-object}
\end{subfigure}
\hfill
\begin{subfigure}[b]{0.49\linewidth}
    \centering
    \includegraphics[width=\linewidth]{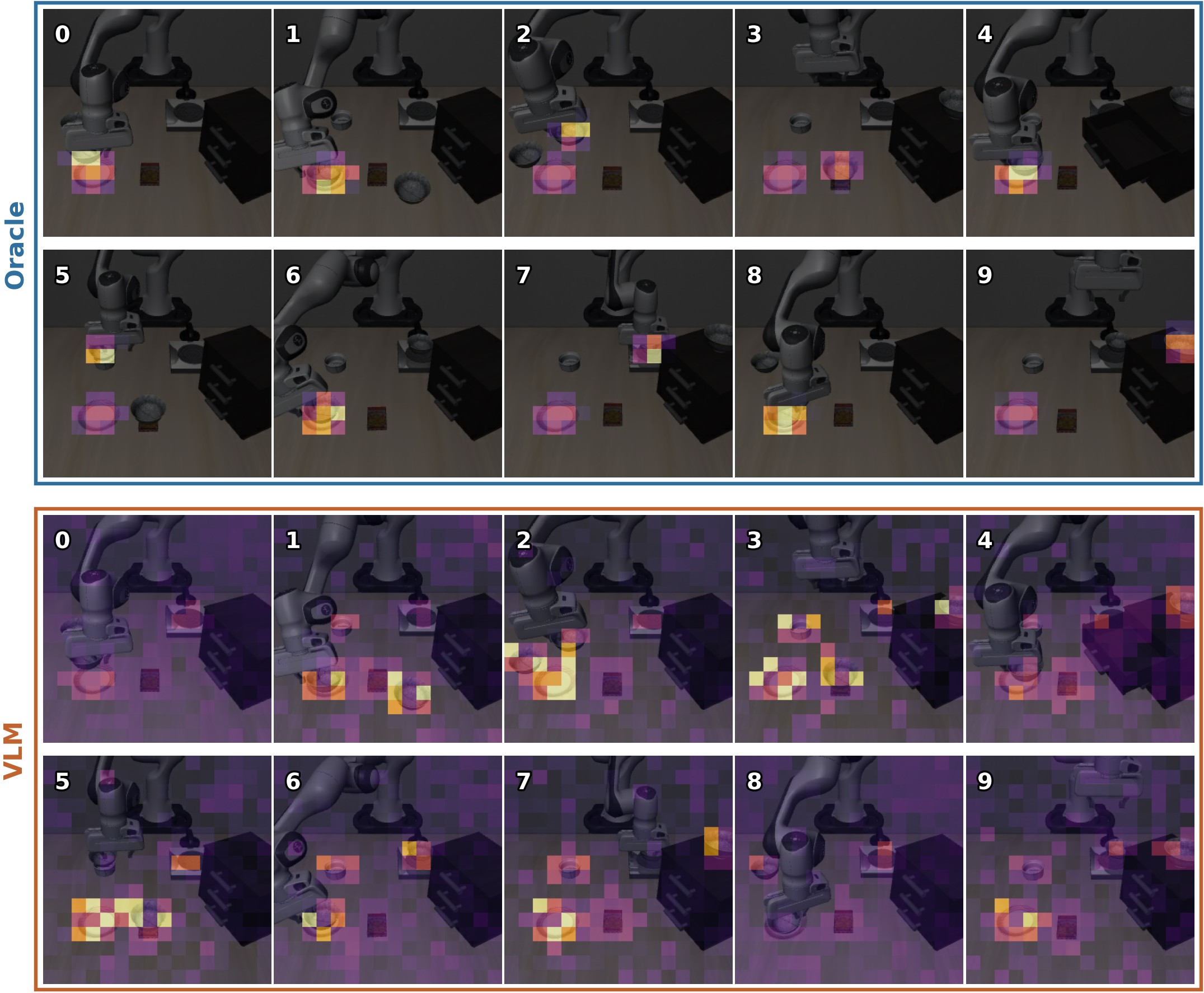}
    \caption{LIBERO-Spatial}
    \label{fig:appendix-heatmaps-spatial}
\end{subfigure}

\vspace{4pt}

\begin{subfigure}[b]{0.49\linewidth}
    \centering
    \includegraphics[width=\linewidth]{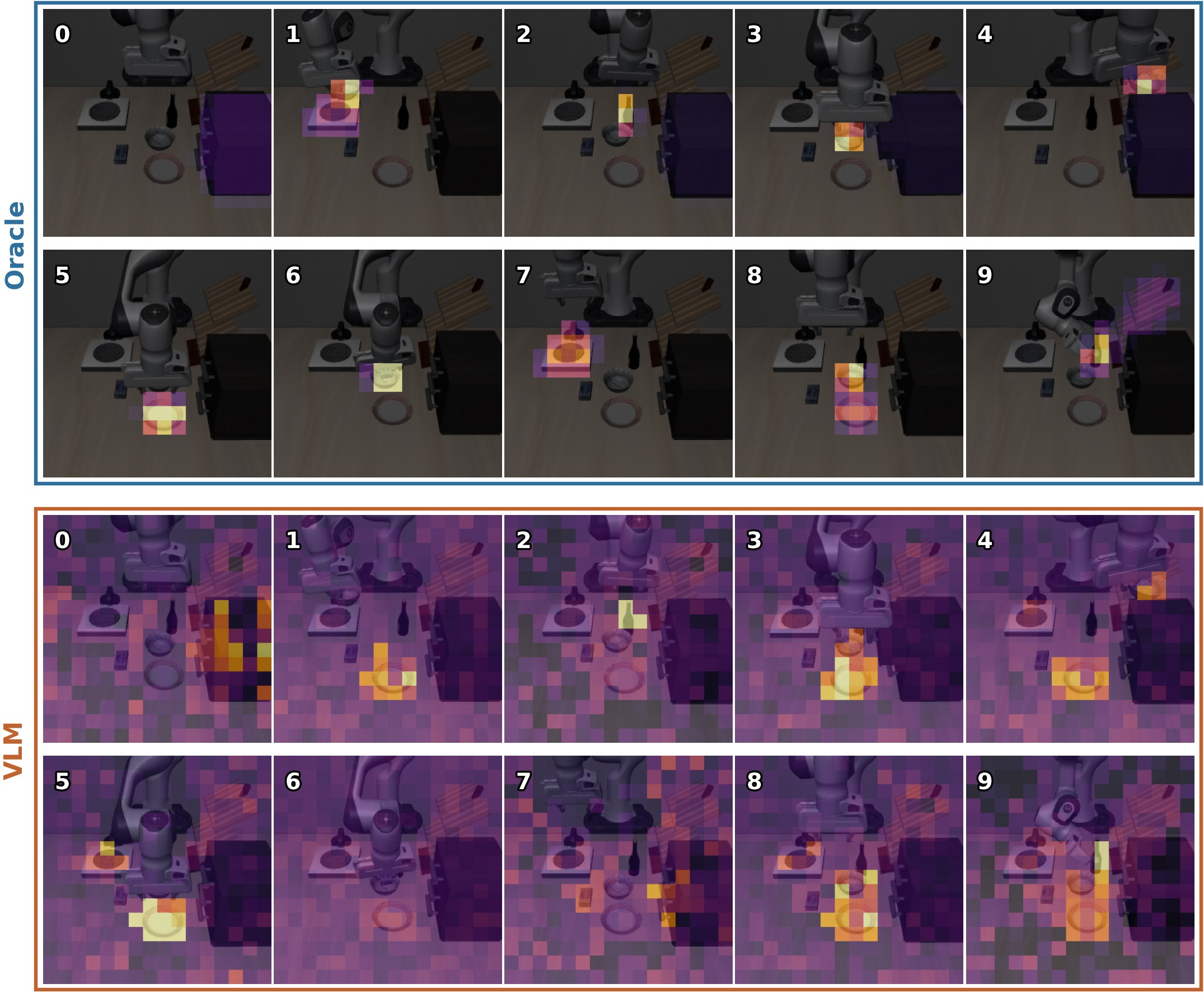}
    \caption{LIBERO-Goal}
    \label{fig:appendix-heatmaps-goal}
\end{subfigure}
\hfill
\begin{subfigure}[b]{0.49\linewidth}
    \centering
    \includegraphics[width=\linewidth]{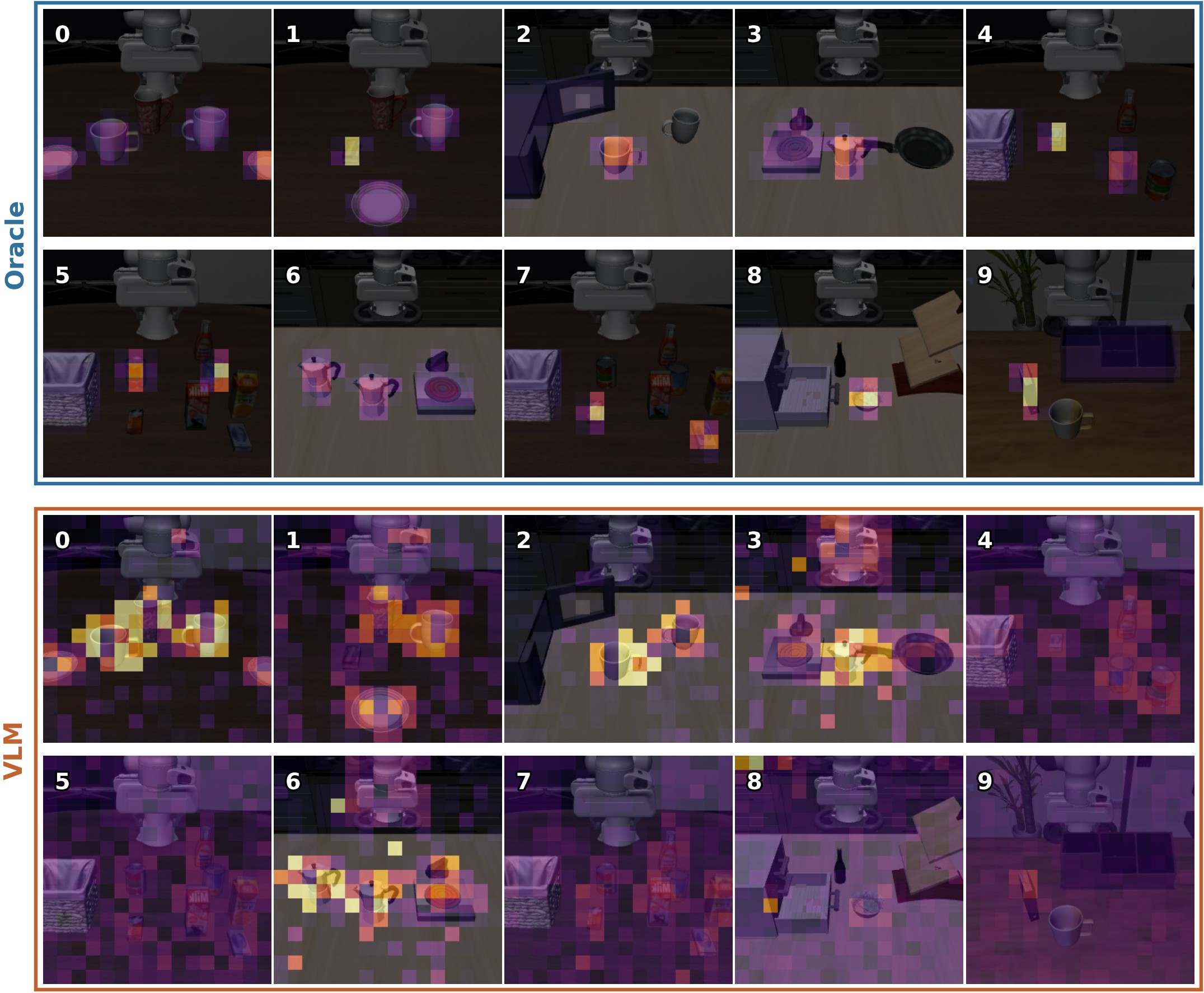}
    \caption{LIBERO-Long}
    \label{fig:appendix-heatmaps-long}
\end{subfigure}
\vspace{-4pt}
\caption{\textbf{Oracle masks against VLM localization heatmaps}. One randomly
sampled training frame per task, numbered 0 to 9 within each block, following the
task order of Tables~\ref{tab:appendix-prompts-a}
and~\ref{tab:appendix-prompts-b}. In every subfigure, the upper block shows the
ground-truth oracle mask and the lower block the VLM similarity heatmap produced
by a trained heatmap-conditioned \ourvla checkpoint on the identical frames,
captured with a forward hook on the projection layer of the \textit{where}
module, \ie the exact tensor that the policy is conditioned on. Each block is
independently color-normalized to its own median per-task peak, because the VLM
signal is one to two orders of magnitude weaker in absolute mass and a shared
scale would render the lower blocks blank. Intensity should therefore be compared
within a block and not across blocks.}
\label{fig:appendix-heatmaps}
\vspace{-12pt}
\end{figure*}

\noindent\textbf{Conditioning the \textit{where} module with an oracle
localizer}: The localization heatmaps that normally drive the \textit{where}
module (Sec.~\ref{ssec:architecture}) carry almost no signal in LIBERO.
Fig.~\ref{fig:appendix-heatmaps} compares them, for every suite, against a
ground-truth oracle mask computed on identical frames.
The oracle mask localizes the target(s) (\ie areas of interest --- both the objects to be manipulated but also the target locations) correctly, whereas the MetaCLIP2 similarity
map is either close to uniform or focuses on the wrong area.
Audited over full policy-driven rollouts, its arg-max patch coincides with the
target in \textbf{0 of 3{,}775} steps across the ten LIBERO-Object tasks, and it
does not even lock onto a consistent distractor (the robot arm accounts for only
5\% of them).
We attribute this to the renderer rather than to training.
LIBERO's objects are small, low-texture, near-duplicate meshes rendered at
$256{\times}256$ (three of the ten LIBERO-Object items are literally colored
boxes), which lies far outside the distribution that CLIP-family patch features
were trained on, and both backbones remain frozen, so no amount of policy
training can sharpen them.

We therefore replace the similarity map with the ground-truth footprint of the
task's objects of interest, taken from LIBERO's MuJoCo segmentation buffer,
pooled to the same $16{\times}16$ grid and normalized into a distribution.
The identical extraction code builds the training table and renders the map live
at evaluation, so there is no train/eval skew.
This does not inflate the reported scores, because the localizer is not the
quantity under test.
The \textit{where} module consumes an interface, \ie a distribution over image
patches, and the question asked here is whether the policy built on top of that
interface still scales with tasks and data.
Substituting a perfect producer of that distribution removes the frozen VLM as a
confounding factor, in the same way that one would freeze a perception module in
order to isolate a controller.
Since it consumes privileged simulator state, the resulting numbers are an upper
bound on the contribution of the \textit{where} module rather than a deployable
system, and we flag them as such wherever they appear.
On the real robot the same interface is fed by an actual VLM, and the results of
the main paper are the deployable ones.

\noindent\textbf{Training and evaluation}: We train a single \ourvla from scratch on the 4 LIBERO subsets (\ie object, spatial, goal and long), on a single NVIDIA GeForce RTX 4090.
W train for $35,000$ steps with a batch size of $32$, using AdamW with a peak learning rate of $1 \times 10^{-4}$ cosine-annealed to $1 \times 10^{-5}$,
and DINOv3 ViT-B/16 and MetaCLIP2 ViT-H features computed online.
We use the \textit{where} $\rightarrow$ \textit{what} module sequence, a chunk size of $20$ with $10$ actions executed per inference, and an observation history of $10$ steps.
We also re-train a two \otter models from scratch on LIBERO (one with CLIP and another with MetaCLIP2 as backbone), because in the original paper, no results are reported for LIBERO-Long and no results exist for our update version. 
Again, we train for for $35,000$ steps with a batch size of $32$, using AdamW with a peak learning rate of $1 \times 10^{-4}$ cosine-annealed to $1 \times 10^{-5}$.
Similarly to \ourvla, a chunk size of $20$ with $10$ actions executed per inference, and an observation history of $10$ steps.
Evaluation is $50$ trials per task from LIBERO's canonical initial states, \ie
$2,000$ rollouts per model, with success determined by LIBERO's own predicates.

\begin{table*}[p]
\centering
\footnotesize\setlength{\tabcolsep}{4pt}
\renewcommand{\arraystretch}{0.95}
\begin{tabularx}{\textwidth}{@{}c Y{1.1} Y{1.2} Y{0.7}@{}}
\toprule
\# & Task instruction & \textit{what} (skill) & \textit{where} (localized) \\
\midrule
\multicolumn{4}{@{}l}{\textit{LIBERO-Object}: the skill string is constant, so all discrimination falls to the \textit{where} module.}\\
\midrule
0 & pick up the alphabet soup and place it in the basket & put in the basket & alphabet soup, basket \\
1 & pick up the cream cheese and place it in the basket & put in the basket & cream cheese, basket \\
2 & pick up the salad dressing and place it in the basket & put in the basket & salad dressing, basket \\
3 & pick up the bbq sauce and place it in the basket & put in the basket & bbq sauce, basket \\
4 & pick up the ketchup and place it in the basket & put in the basket & ketchup, basket \\
5 & pick up the tomato sauce and place it in the basket & put in the basket & tomato sauce, basket \\
6 & pick up the butter and place it in the basket & put in the basket & butter, basket \\
7 & pick up the milk and place it in the basket & put in the basket & milk, basket \\
8 & pick up the chocolate pudding and place it in the basket & put in the basket & chocolate pudding, basket \\
9 & pick up the orange juice and place it in the basket & put in the basket & orange juice, basket \\
\midrule
\multicolumn{4}{@{}l}{\textit{LIBERO-Spatial}: the localized entities are constant, so all discrimination falls to the \textit{what} module.}\\
\midrule
0 & \ldots between the plate and the ramekin\ldots & pick up between the plate and the ramekin and place on the plate & bowl, plate \\
1 & \ldots next to the ramekin\ldots & pick up next to the ramekin and place on the plate & bowl, plate \\
2 & \ldots from table center\ldots & pick up at table center and place on the plate & bowl, plate \\
3 & \ldots on the cookie box\ldots & pick up on the cookie box and place on the plate & bowl, plate \\
4 & \ldots in the top drawer of the wooden cabinet\ldots & pick up in the top drawer of the wooden cabinet and place on the plate & bowl, plate \\
5 & \ldots on the ramekin\ldots & pick up on the ramekin and place on the plate & bowl, plate \\
6 & \ldots next to the cookie box\ldots & pick up next to the cookie box and place on the plate & bowl, plate \\
7 & \ldots on the stove\ldots & pick up on the stove and place on the plate & bowl, plate \\
8 & \ldots next to the plate\ldots & pick up next to the plate and place on the plate & bowl, plate \\
9 & \ldots on the wooden cabinet\ldots & pick up on the wooden cabinet and place on the plate & bowl, plate \\
\bottomrule
\end{tabularx}
\vspace{4pt}
\caption{\textbf{Prompt decomposition, LIBERO-Object and LIBERO-Spatial}. Every
LIBERO-Spatial instruction reads ``pick up the black bowl
\textit{$\langle$landmark$\rangle$} and place it on the plate'', and since
several landmarks differ only by a preposition, the skill embeddings are
mean-centered over that suite before use.}
\label{tab:appendix-prompts-a}
\end{table*}

\begin{table*}[p]
\centering
\footnotesize\setlength{\tabcolsep}{4pt}
\renewcommand{\arraystretch}{0.95}
\begin{tabularx}{\textwidth}{@{}c Y{1.1} Y{1.2} Y{0.7}@{}}
\toprule
\# & Task instruction & \textit{what} (skill) & \textit{where} (localized) \\
\midrule
\multicolumn{4}{@{}l}{\textit{LIBERO-Goal}: the only suite in which both slots vary per task.}\\
\midrule
0 & open the middle drawer of the cabinet & open the middle drawer & cabinet middle drawer \\
1 & put the bowl on the stove & put on the stove & bowl, stove cook region \\
2 & put the wine bottle on top of the cabinet & put on top of the cabinet & wine bottle, cabinet top \\
3 & open the top drawer and put the bowl inside & open the top drawer $/$ put inside & bowl, cabinet top drawer \\
4 & put the bowl on top of the cabinet & put on top of the cabinet & bowl, cabinet top \\
5 & push the plate to the front of the stove & push to the front of the stove & plate \\
6 & put the cream cheese in the bowl & put in the bowl & cream cheese, bowl \\
7 & turn on the stove & turn on & stove \\
8 & put the bowl on the plate & put on the plate & bowl, plate \\
9 & put the wine bottle on the rack & put on the rack & wine bottle, wine rack \\
\midrule
\multicolumn{4}{@{}l}{\textit{LIBERO-Long}: multi-step tasks, in which both slots may hold several concepts.}\\
\midrule
0 & put both the alphabet soup and the tomato sauce in the basket & put the alphabet soup and tomato sauce in the basket & alphabet soup, tomato sauce, basket \\
1 & put both the cream cheese box and the butter in the basket & put the cream cheese box and butter in the basket & cream cheese, butter, basket \\
2 & turn on the stove and put the moka pot on it & turn on the stove $/$ put one pot on the stove & moka pot, stove \\
3 & put the black bowl in the bottom drawer of the cabinet and close it & put in the bottom drawer $/$ close the drawer & bowl, white cabinet \\
4 & put the white mug on the left plate and put the yellow and white mug on the right plate & put on the left plate $/$ put on the right plate & both mugs, both plates \\
5 & pick up the book and place it in the back compartment of the caddy & put in the back compartment of the caddy & book, desk caddy \\
6 & put the white mug on the plate and put the chocolate pudding to the right of the plate & put on the plate $/$ put to the right of the plate & mug, chocolate pudding, plate \\
7 & put both the alphabet soup and the cream cheese box in the basket & put the alphabet soup and cream cheese box in the basket & alphabet soup, cream cheese, basket \\
8 & put both moka pots on the stove & put both pots on the stove & both moka pots, stove \\
9 & put the yellow and white mug in the microwave and close it & put in the microwave $/$ close the microwave & mug, microwave \\
\bottomrule
\end{tabularx}
\vspace{4pt}
\caption{\textbf{Prompt decomposition, LIBERO-Goal and LIBERO-Long}. A ``$/$''
marks a task that chains two verbs, in which case the two phrases are embedded
separately and summed. The destination of LIBERO-Goal task 5
(\texttt{main\_table\_stove\_front\_region}) is a table region with no geometry to
segment, so only the plate is localized. In the three LIBERO-Long tasks that
share the basket destination, the item names are repeated inside the skill phrase
to keep the skill embeddings distinguishable.}
\label{tab:appendix-prompts-b}
\end{table*}

\noindent\textbf{Prompt decomposition}:
Tables~\ref{tab:appendix-prompts-a} and~\ref{tab:appendix-prompts-b} list, for
all 40 tasks, the skill string that conditions the \textit{what} module and the
objects that the oracle localizes for the \textit{where} module, as generated by a high-level VLM planner (\ie Opus 5.1).
The two slots are kept disjoint by construction, in that the verb and the
destination are routed to the \textit{what} module while the entities to be
localized are routed to the \textit{where} module, so that the \textit{what}
module cannot solve localization by memorizing skill-object pairings.
Object targets are read directly from the \texttt{obj\_of\_interest} declaration
of each task's BDDL file, which lists the manipulated object together with its
destination, so most tasks localize two entities at once.
Each entity is normalized to unit mass before summing, as otherwise a large
destination (the basket, at roughly $3000$\,px against a $365$\,px soup can)
would occupy roughly $89\%$ of the map and render the target numerically
invisible.

\begin{table*}[t]
\centering
\small\setlength{\tabcolsep}{2pt}
\renewcommand{\arraystretch}{0.95}
\caption{\textbf{LIBERO success rate (\%)}, for a single policy trained on all four suites (40 tasks, in total) and evaluated with 50 trials per
task from the canonical initial states. 
\textit{Params} counts all model trainable parameters.
$\dagger$ indicates results taken from~\cite{huang2025otter} for \otter trained and evaluated on LIBERO's object, spatial and goal, while the \otter (CLIP) entry corresponds to our training of \otter on all four tasks (\ie +long). $\ddagger$ indicates results taken from the \href{https://huggingface.co/docs/lerobot/pi05}{LeRobot LIBERO evaluation page}, for training and evaluating \pizerofive~\cite{black2025pi05} on LIBERO's object, spatial, goal and long. 
}
\label{tab:appendix-libero-results}
\vspace{2pt}
\begin{tabular*}{\textwidth}{@{\extracolsep{\fill}}l r ccccc@{}}
\toprule
Method & Params & Object & Spatial & Goal & Long & Average \\
\midrule
\otter (CLIP)$^\dagger$ & --- & 89.0 & 84.0 & 79.0 & n/a & 84.0 \\
\otter (CLIP) & 45M & 97.0 & 74.8 & 80.2 & 32.6 & 71.2 \\
\otter (MetaCLIP2) & 67M & 98.8 & 73.2 & 88.0 & 48.6 & 77.2 \\
\pizerofive$^\ddagger$  & 3.6B & 99.0 & 97.0 & 98.0 & 96.0 & 97.5 \\
\midrule
\ourvla (ours) & 56M & 99.8 & 90.6 & 95.4 & 68.4 & 88.6 \\
\bottomrule
\end{tabular*}
\end{table*}

\begin{table*}[ht]
\centering
\vspace{5pt}
\caption{\textbf{LIBERO-Plus performance}. A single policy trained on all 40 LIBERO tasks, evaluated on canonical states (\textit{Original}) and along seven perturbation axes, with the \textit{drop from Original} shaded below each row. 
}
\label{tab:libero_plus_appendix}
\vspace{-4pt}
\small\setlength{\tabcolsep}{4pt}
\renewcommand{\arraystretch}{0.85}
\begin{tabular*}{0.95\textwidth}{@{\extracolsep{\fill}}lc||ccccccc||r@{}}
\toprule
Method & Original & Camera & Robot & Lang. & Light & BG & Noise & Layout & Total \\
\midrule
\midrule
OpenVLA & 76.5 & 1.1 & 4.1 & 26.8 & 4.4 & 25.3 & 19.3 & 31.6 & 16.0 \\
& &\dcell{75.4} &\dcell{72.4} &\dcell{49.7} &\dcell{72.1} &\dcell{51.2} &\dcell{57.2} &\dcell{44.9} &\dcell{60.5} \\
\midrule
OpenVLA-OFT & 97.1 & 59.7 & 37.2 & 81.5 & 85.8 & 92.4 & 76.7 & 77.1 & 71.4 \\
& &\dcell{37.4} &\dcell{59.9} &\dcell{15.6} &\dcell{11.3} &\dcell{4.7} &\dcell{20.4} &\dcell{20.0} &\dcell{25.7} \\
\midrule
OpenVLA-OFT\_w & 95.3 & 16.8 & 43.7 & 73.2 & 68.2 & 92.5 & 51.4 & 72.3 & 57.5 \\
& &\dcell{78.5} &\dcell{51.6} &\dcell{22.1} &\dcell{27.1} &\dcell{2.8} &\dcell{43.9} &\dcell{23.0} &\dcell{37.8} \\
\midrule
OpenVLA-OFT\_m & 97.6 & 57.9 & 30.6 & 83.6 & 91.6 & 83.6 & 76.3 & 73.2 & 69.5 \\
& &\dcell{39.7} &\dcell{67.0} &\dcell{14.0} &\dcell{6.0} &\dcell{14.0} &\dcell{21.3} &\dcell{24.4} &\dcell{28.1} \\
\midrule
$\pi_0$ & 94.2 & 15.8 & 6.6 & 61.0 & 79.6 & 78.5 & 79.4 & 70.4 & 53.8 \\
& &\dcell{78.4} &\dcell{87.6} &\dcell{33.2} &\dcell{14.6} &\dcell{15.7} &\dcell{14.8} &\dcell{23.8} &\dcell{40.4} \\
\midrule
$\pi_0$-fast & 85.5 & 66.4 & 24.8 & 63.3 & 73.0 & 67.7 & 75.8 & 70.3 & 62.5 \\
& &\dcell{19.1} &\dcell{60.7} &\dcell{22.2} &\dcell{12.5} &\dcell{17.8} &\dcell{9.7} &\dcell{15.2} &\dcell{23.0} \\
\midrule
Nora & 87.9 & 4.0 & 41.1 & 67.0 & 31.0 & 50.5 & 17.6 & 63.9 & 38.7 \\
& &\dcell{83.9} &\dcell{46.8} &\dcell{20.9} &\dcell{56.9} &\dcell{37.4} &\dcell{70.3} &\dcell{24.0} &\dcell{49.2} \\
\midrule
WorldVLA & 79.1 & 0.3 & 30.2 & 44.2 & 29.4 & 14.5 & 12.2 & 39.4 & 24.3 \\
& &\dcell{78.8} &\dcell{48.9} &\dcell{34.9} &\dcell{49.7} &\dcell{64.6} &\dcell{66.9} &\dcell{39.7} &\dcell{54.8} \\
\midrule
UniVLA & 95.2 & 4.3 & 50.3 & 71.8 & 59.1 & 80.0 & 25.3 & 34.3 & 44.0 \\
& &\dcell{90.9} &\dcell{44.9} &\dcell{23.4} &\dcell{36.1} &\dcell{15.2} &\dcell{69.9} &\dcell{60.9} &\dcell{51.2} \\
\midrule
RIPT-VLA & 97.5 & 58.3 & 36.7 & 80.1 & 87.9 & 90.4 & 73.8 & 76.5 & 70.4 \\
& &\dcell{39.2} &\dcell{60.8} &\dcell{17.4} &\dcell{9.6} &\dcell{7.1} &\dcell{23.7} &\dcell{21.0} &\dcell{27.1} \\
\midrule
\midrule
\pizerofivebold (from~\cite{han2026geometricactionmodelrobot}) & 96.9 & \textbf{72.0} & \textbf{76.6} & 86.5 & \textbf{96.1} & \textbf{95.2} & 86.7 & \textbf{86.0} & \textbf{84.6} \\
& &\dcell{24.9} &\dcell{20.3} &\dcell{10.4} &\dcell{0.8} &\dcell{1.7} &\dcell{10.2} &\dcell{10.9} &\dcell{12.3} \\
\midrule
\otter (CLIP) & 71.2 & 43.2 & 47.6 & 15.0 & 66.5 & 65.7 & 63.8 & 53.8 & 50.8 \\
& &\dcell{28.0} &\dcell{23.5} &\dcell{56.1} &\dcell{4.6} &\dcell{5.4} &\dcell{7.4} &\dcell{17.4} &\dcell{20.4} \\
\midrule
\otter (MetaCLIP2) & 77.2 & 49.1 & 47.6 & 20.6 & 70.8 & 68.6 & 60.4 & 52.3 & 51.4 \\
& &\dcell{28.1} &\dcell{29.6} &\dcell{56.6} &\dcell{6.4} &\dcell{8.6} &\dcell{16.8} &\dcell{24.9} &\dcell{25.8} \\
\midrule
\ourvla  & 88.6 & 9.8 & 59.9 & \textbf{88.6} & 87.0 & 87.5 & \textbf{88.1} & 67.3 & 69.7 \\
& &\dcell{78.8} &\dcell{28.7} &\dcell{0.0} &\dcell{1.6} &\dcell{1.1} &\dcell{0.5} &\dcell{21.3} &\dcell{18.8} \\
\bottomrule
\end{tabular*}
\end{table*}



\noindent\textbf{Results}: Table~\ref{tab:appendix-libero-results} reports the pooled 40-task policies.
\ourvla reaches $88.6\%$ on average with $56$M parameters, showcasing that our design preserves traditional VLA capabilities, \ie ability to handle dataset scaling and solve multitask problems. 
\otter, even with MetaCLIP2 as a backbone, seems to fail to perform well in long-horizon tasks, which significantly hurts its overall performance, scoring a success rate of $77.2\%$.
The highest success rate is achieved by \pizerofive, scoring $97.5\%$. 
However, this is achieved with roughly $\times100$ more trainable parameters, and a very costly pre-training. 
Consequently, even though \ourvla is significantly more compact and trained from scratch, it still performs competitively on LIBERO, which highlights that the modular design that unlocks compositional generalization \textit{does not} sacrifice general VLA capabilities (\ie multitask applications and dataset scaling).

\subsection{LIBERO-Plus}
\label{sec:appendix-libero-plus}

Success on LIBERO's canonical initial states says little about how a policy behaves once the scene changes. LIBERO-Plus covers this by perturbing the same four suites along seven axes: camera pose, robot initial configuration, language phrasing, lighting, background, sensor noise and object layout. We evaluate the same 40-task checkpoint of Sec.~\ref{ssec:sim}, with no retraining or fine-tuning. We compare it against the models evaluated by the benchmark authors, $\pi_{0.5}$, and our two OTTER re-implementations. As in Sec.~\ref{ssec:sim}, the \textit{where} module uses the oracle localizer. The w$^2$VLA numbers are therefore an upper bound on the contribution of the \textit{where} module, not a deployable system.

\textbf{Overall results.} Table~\ref{tab:libero_plus_appendix} shows that w$^2$VLA reaches a total success rate of 69.7\%. This places it fourth of the fourteen evaluated models, behind $\pi_{0.5}$ (84.6\%), OpenVLA-OFT (71.4\%) and RIPT-VLA (70.4\%), and level with OpenVLA-OFT$_m$ (69.5\%). All of these carry billions of parameters and large-scale robot pre-training, whereas w$^2$VLA has 56M parameters trained from scratch. The drop from the unperturbed setting is more informative than the absolute score. It is 18.8 points for w$^2$VLA, the second smallest in the table after $\pi_{0.5}$ (12.3). The next smallest are OTTER (CLIP) at 20.4, $\pi_0$-FAST at 23.0 and OpenVLA-OFT at 25.7, while $\pi_0$, UniVLA, NORA, WorldVLA and OpenVLA lose between 40 and 61 points. Against OTTER, the closest baseline in size and design (a frozen VLM backbone and a compact action expert), w$^2$VLA is 18.3 points higher in total. It also degrades by 7.0 points less than the MetaCLIP2 variant, even though both share the same backbone. \textit{The robustness gap therefore comes from how the VLM signal is routed into the policy, not from the VLM itself}.

\textbf{Appearance perturbations.} The advantage of w$^2$VLA is concentrated on the axes that change how the scene looks but not its geometry. Lighting, background and sensor noise cost w$^2$VLA only 1.6, 1.1 and 0.5 points. Its drops under background and sensor noise are the smallest in the table, and its lighting drop is second only to $\pi_{0.5}$ (0.8). Under sensor noise, w$^2$VLA obtains the highest absolute score of all models (88.1\%), ahead of $\pi_{0.5}$ (86.7\%, $\downarrow$10.2), while $\pi_0$ and OpenVLA-OFT lose 14.8 and 20.4 points. This matches the design of the architecture. Random patch masking (Sec.~\ref{sec:methodology}) pushes the policy to rely on the \textit{where} and \textit{what} conditioning signals rather than on the raw visual stream. Both signals are, by construction, unaffected by how the scene is lit, textured or corrupted.

\textbf{Robot initialization.} The same reasoning extends to the robot initial configuration axis. Here w$^2$VLA retains 59.9\%, second only to $\pi_{0.5}$ (76.6\%) and well above every OpenVLA-OFT variant ($\leq$43.7\%) and $\pi_0$ (6.6\%). The proprioceptive history is modulated by task information that does not depend on the arm's starting pose, so an unfamiliar initial configuration does not change what the policy is conditioned to do.

\textbf{Language perturbations.} w$^2$VLA shows no degradation under paraphrasing and achieves the highest score on this axis (88.6\%). By contrast, $\pi_{0.5}$ loses 10.4 points and both OTTER variants collapse by more than 56 points. This number should be read as a property of the interface rather than of the low-level policy. The \textit{what} and \textit{where} modules consume the decomposed skill and object strings of Tables~\ref{tab:appendix-prompts-a} and~\ref{tab:appendix-prompts-b}. Rephrasings of the raw instruction are therefore absorbed by the high-level planner and never reach the action expert. The burden of understanding free-form language falls on the decomposition stage, exactly as in the hierarchical setting of Sec.~\ref{sec:methodology}.

\textbf{Object layout.} Layout perturbations reduce w$^2$VLA to 67.3\% ($\downarrow$21.3), a mid-table result. $\pi_{0.5}$ (86.0\%), OpenVLA-OFT (77.1\%) and RIPT-VLA (76.5\%) all score higher. Because the oracle localizer still marks the relocated targets correctly, this drop cannot be attributed to localization. It reflects the action expert's limited ability to adapt its trajectories to object placements outside the training distribution. This is the a limitation that larger and more diverse robot datasets are expected to reduce.

\textbf{Camera perturbations.} Camera pose is the one axis on which w$^2$VLA fails, dropping to 9.8\% ($\downarrow$78.8). The \textit{where} module conditions the policy on a distribution over image patches, and the policy learns a mapping from that 2D grid to end-effector motion that holds only under the training camera extrinsics. Once the viewpoint moves, the oracle still highlights the correct patches, but the same patch location no longer corresponds to the same point in the workspace. The spatial conditioning then becomes misleading rather than merely uninformative. Several large VLAs are equally fragile on this axis ($\pi_0$ 15.8\%, UniVLA 4.3\%, NORA 4.0\%, OpenVLA 1.1\%). OTTER, however, retains 43--49\% despite using the same frozen VLM features, which shows that the failure is specific to conditioning explicitly on a 2D spatial grid. Excluding this axis, the unweighted mean of w$^2$VLA over the remaining six axes is 78.1\%, second only to $\pi_{0.5}$ (87.9\%) and ahead of OpenVLA-OFT (75.1\%) and RIPT-VLA (74.2\%). The most direct fix is to ground \textit{where} in 3D, for example by lifting the heatmaps into the robot base frame with depth and camera calibration, which would make the signal viewpoint-invariant. A complementary option is to pre-train the action expert on OXE with the VLM encoders kept frozen, supplying the viewpoint and embodiment diversity that benefits $\pi_{0.5}$.

\textbf{Summary.} A compact, compositional policy trained from scratch keeps its canonical-state performance under appearance, language and robot-initialization shifts better than most billion-parameter VLAs. Its weak points are clearly identifiable: viewpoint changes, which follow from the 2D \textit{where} interface, and layout changes, which follow from limited trajectory diversity. Neither is a limitation of the compositional decomposition itself.


\end{document}